\pdfoutput=1
\documentclass[11pt]{article}

\usepackage{coling}

\usepackage{times}
\usepackage{latexsym}

\usepackage[T1]{fontenc}
\usepackage[utf8]{inputenc}

\usepackage{microtype}

\usepackage{inconsolata}

\usepackage{graphicx}

\title{InkubaLM: A small language model for low-resource African languages}

\author{\normalsize Atnafu Lambebo Tonja$^{1,2,*}$,  Bonaventure F. P. Dossou$^{1,3,*}$, Jessica Ojo$^{1,3}$, \\ 
\textbf{\normalsize  Jenalea Rajab$^{1,5}$, Fadel Thior$^{1}$, Eric Peter Wairagala$^{1}$, Anuoluwapo Aremu$^{1}$}, \\
\textbf{\normalsize Pelonomi Moiloa$^{1}$, Jade Abbott$^{1}$, Vukosi Marivate$^{1,4}$, Benjamin Rosman$^{1,5}$} \\\\
\footnotesize
$^1$ Lelapa AI, $^2$ MBZUAI, $^3$ McGill University \& Mila Quebec AI Institute, \\
\footnotesize
$^4$ DSFSI - University of Pretoria, $^5$ RAIL Lab - University of the Witwatersrand
\\
\footnotesize
$^*$ Equal Contributions
\\
}

\begin{document}
\maketitle
\begin{abstract}
High-resource language models often fall short in the African context, where there is a critical need for models that are efficient, accessible, and locally relevant, even amidst significant computing and data constraints. This paper introduces InkubaLM, a small language model with 0.4 billion parameters, which achieves performance comparable to models with significantly larger parameter counts and more extensive training data on tasks such as machine translation, question-answering, AfriMMLU, and the AfriXnli task. Notably, InkubaLM outperforms many larger models in sentiment analysis and demonstrates remarkable consistency across multiple languages. This work represents a pivotal advancement in challenging the conventional paradigm that effective language models must rely on substantial resources. Our model and datasets are publicly available \footnote{\url{https://huggingface.co/lelapa}} to encourage research and development on low-resource languages.
\end{abstract}

\section{Introduction}
The field of Natural Language Processing (NLP) has witnessed transformative growth with the advent of Large Language Models (LLMs). These models, characterized by billions of parameters, have set new benchmarks in tasks such as language translation, sentiment analysis, and more sophisticated applications like creative writing and conversational AI. Notable examples include models like GPT-3 \cite{GPT} and BERT \cite{BERT}, which have influenced academic research and have seen extensive adoption across industries. These models are central to enhancing user interactions and automating content generation, leading to widespread use in consumer applications like chatbots, virtual assistants, and recommendation systems.

LLMs have predominantly been developed and trained on high-resource languages with extensive datasets \cite{minaee2024largelanguagemodelssurvey}, such as English, Chinese, and Spanish, leaving low-resource languages, particularly in Africa, at a significant disadvantage. The main challenge is the scarcity of quality textual data, as LLMs require vast data to train effectively. This data is often fragmented, non-standardized, or non-existent for low-resource languages, and the absence of essential tools like part-of-speech taggers or annotated datasets limits progress. With its over 2,000 languages, Africa exemplifies this challenge, as many of these languages are underrepresented in digital resources and NLP research \cite{Nekoto}. Additionally, the computational resources required to train and deploy LLMs are frequently scarce in many African regions, limiting the ability of researchers and developers to leverage these powerful tools for use within their communities \cite{masakhane}.

Developing models that are easy to refine, fine-tune, explore, and deploy in cost-effective ways on limited hardware is essential. While open-source models have made strides in bridging the language gap, more efforts are needed to create models that are not only efficient but also locally relevant. InkubaLM, which can move 250 times its weight, inspired by the strength and adaptability of the dung beetle, exemplifies this approach by offering a smaller yet powerful model designed to empower African communities. Accompanied by two datasets, InkubaLM represents a pioneering initiative to distribute the computational load and enable access to NLP tools such as Machine Translation, Sentiment Analysis, Named Entity Recognition (NER), Parts of Speech Tagging (POS), Question Answering, and Topic Classification for their languages.

Our contributions are as follows: \textbf{(1)} we introduce InkubaLM, the first open-source small multilingual language model for African languages, and \textbf{(2)} we introduce the first instruction dataset for five NLP tasks in African languages. \textbf{(3)} we are also releasing the monolingual dataset used for training in 5 African languages to encourage further research on low-resource languages.

\section{Related Work}

\subsection{Low-resource Languages}

Given the vast linguistic diversity and the scarcity of digital resources for many languages, particularly in Africa, researchers have been exploring various approaches to make LLMs more inclusive and effective for these languages.

One approach to addressing the challenges in low-resource settings involves multilingual models and cross-lingual transfer learning. Models like Multilingual BERT (mBERT) \cite{BERT}, XLM-R \cite{XLM-R}, and Llama 3 \cite{dubey2024llama3herdmodels} have been trained on data from multiple languages, including some low-resource languages. These models can leverage shared representations across languages, allowing for improved performance even in cases where labeled data is scarce. However, the effectiveness of these models in truly low-resource languages remains limited due to the small amount of training data available for these languages.

In addition to general-purpose multilingual models \cite{qin2024multilinguallargelanguagemodel}, there has been a push toward developing specialized models tailored to specific low-resource languages \cite{hedderich2021surveyrecentapproachesnatural}. These models often utilize transfer learning techniques, where a pre-trained model on a high-resource language is fine-tuned on a smaller dataset from a low-resource language \cite{Nekoto}. This approach improves the model's performance and reduces the computational resources required, making it more feasible for use in resource-constrained environments.

Despite these advancements, significant challenges remain in making LLMs effective for low-resource languages. Issues such as linguistic bias, model interpretability, and the ethical implications of deploying these models in diverse cultural contexts remain concerning. This paper focuses on developing a culturally aware, efficient, low-resource model that can operate effectively despite limited data and computational power.

\subsection{Small language models}

\citet{zhang2024tinyllama} introduced TinyLlama, a 1.1 billion parameter language model pre-trained on 1 trillion tokens. Despite its compact size, TinyLlama leverages techniques such as FlashAttention to deliver strong performance across various tasks, outperforming many models within its class. Building on this, the authors developed TinyLlama v1.1, which includes specialized versions of models tailored for math, code, and Chinese language tasks, showcasing improved results through a multi-stage pretraining process.

In a related effort to optimize model efficiency, the OneBit framework \cite{xu2024onebit} marks a significant advancement in the quantization of large language models (LLMs) to 1-bit representations. This method drastically reduces computational and memory requirements, enabling the deployment of LLMs on devices with limited resources. Departing from traditional quantization techniques that use 4-bit or 8-bit compression, OneBit achieves an impressive compression ratio while maintaining a balanced trade-off between size reduction and model accuracy across various tasks.

Moving towards resource-efficient training, Inheritune \cite{sanyal2024pre} develops smaller base language models by inheriting layers from a more extensive reference model and training on a significantly reduced dataset. This approach was exemplified using a 1.5 billion parameter model derived from a more significant 3 billion parameter model. Despite training on only 1 billion tokens—just 0.1\% of the original dataset—the resulting model performed comparably to others trained on significantly larger datasets, highlighting its effectiveness in low-data regimes.

Focusing on on-device processing, MobiLlama \cite{thawakar2024mobillama} is a 0.5 billion parameter small language model (SLM) optimized explicitly for resource-constrained devices. MobiLlama is designed for energy efficiency, low memory usage, and faster inference times, making it ideal for on-device applications. The researchers employed a parameter-sharing technique across transformer layers, enabling the model to retain high accuracy while minimizing training and deployment costs. MobiLlama was evaluated across nine benchmarks, consistently outperforming comparable models, especially in efficiency on low-end hardware.

Regarding model performance in specific tasks, \citet{lepagnol2024small} conducted a study on small language models (SLMs) in zero-shot text classification. Their research involved testing models ranging from 77M to 40B parameters across 15 diverse datasets. The results showed that these smaller models not only match but sometimes surpass the performance of their larger counterparts, creating an open-source repository documenting their methodologies.

\citet{scaria2024can} investigated the capacity of small language models to learn, retain, and unlearn noise patterns. Their study involved models like Olmo 1B, Qwen1.5 1.8B, Gemma 2B, and Phi2 2.7B. It revealed that while these models could learn and even eliminate noise, their performance varied significantly depending on the type of noise introduced, particularly at the character level.

\citet{zhu2024llava} introduced LLaVA-Phi, an efficient multi-modal assistant that harnesses the power of the small language model Phi-2 to facilitate multi-modal dialogues. Despite having only 2.7 billion parameters, LLaVA-Phi demonstrated commendable performance across benchmarks, including visual comprehension and reasoning tasks. The model opens new avenues for applications in time-sensitive environments and systems requiring real-time interaction, proving the potential of smaller language models in sophisticated tasks while maintaining greater resource efficiency.

For natural language processing tasks, \citet{brei2024leveraging} addressed the challenge of translating natural language into SPARQL queries using SLMs. Models such as BART and M2M100 were employed across datasets like QALD and CoyPu, achieving solid results in SPARQL translation, though T5 models struggled with accuracy.

In another approach to model efficiency, \citet{song2024achieving} focused on achieving sparse activation in SLMs. They developed a new attribution metric to overcome the limitations of existing sparse activation techniques, successfully achieving an 80\% sparsification ratio with minimal accuracy loss comparable to larger models.

Lastly, in speech synthesis, \citet{lemerle2024small} introduced the Small-E model, a compact language model enhanced with linear attention. Their work set a new benchmark in zero-shot voice cloning, demonstrating the strong capabilities of small models in this specialized area.
\subsection{LLMs for African languages} 
 
Recent research on LLMs for African languages focuses on several key aspects:  the creation of linguistic resources, the development of adapted models, and the improvement of the performance of LLMs for these languages. \citet{adelani2024irokobenchnewbenchmarkafrican} introduce IrokoBench, a benchmark dataset for 16 African languages, which uncovers significant performance gaps between high-resource and low-resource languages and underscores the importance of developing LLMs tailored to African languages, especially given the disparity between open and proprietary models. \citet{ogueji-etal-2021-small} present a novel approach with AfriBERTa, a multilingual language model specifically trained on low-resource African languages, showing that ``small data'' approaches can outperform traditional models like mBERT and XLM-R without relying on high-resource languages. Building on the challenges of low-resource languages, \citet{joshua2024improving} address the hallucination issues in LLMs, particularly GPT-3.5 turbo, when processing Yoruba by employing Retrieval-Augmented Generation (RAG) techniques, significantly enhancing the models' accuracy and cultural relevance. Similarly, \citet{lawal2024contextual} explore the performance of LLMs in educational settings, particularly in understanding and generating Yoruba primary education science content, and reveal a need for more targeted language-specific models due to the underperformance of existing LLMs in this context. Focusing on another low-resource language, \citet{azime2024enhancing} enhance the LLAMA-2-Amharic model by integrating task-specific and generative datasets, demonstrating improved performance in various NLP tasks and contributing to the growing body of resources for low-resource languages.

\section{Languages} \label{langs}
 Over 3,000 languages are spoken in Africa, a continent known for its rich cultural diversity. From this language, Swahili, Hausa, and Yoruba are widely spoken languages in Africa, with over 218 million speakers. In contrast, isiZulu and isiXhosa are widely spoken languages in South Africa, with over 22 million speakers. We focus on these five languages due to the availability of corpora and because they are the top widely spoken  African languages.

\subsection{Hausa}
The Hausa language is spoken in Northern Nigeria and parts of Cameroon, Chad, Ivory Coast, Ghana, Benin, Togo, Sudan and Niger. The language belongs to the Chadic branch of the Afro-Asiatic language family, boasting about 90M first and second-language speakers. Hausa has between 23-25 consonants and 10 vowels. It also heavily loan words from Arabic and has several dialectal variants.

\subsection{isiZulu}
isiZulu is predominantly spoken in South Africa and some parts of Zimbabwe and has about 14M first language speakers. The language belongs to the Nguni branch of the Niger-Congo language family, with 51 consonants, five vowels, and 15 click sounds. isiZulu also uses three level tones: low, mid, and high tones, which are distinctive in the language. 

\subsection{isiXhosa}
isiXhosa is spoken predominantly in South Africa, parts of Zimbabwe, and Lesotho. It is classified under the Nguni Languages branch of the Niger-Congo language family. The language has 8M native speakers and about 11M second language speakers in South African areas of Eastern Cape, Western Cape, Northern Cape, and Gauteng. isiXhosa is the second most spoken Bantu language in South Africa. The language has 58 consonants, which include 18 click consonants, ten vowels, and two tones, which are rarely marked and use the Latin script.

\subsection{Swahili}
Swahili, also known as KiSwahili, is the most widely spoken language on the African continent, with about 150 million first- and second-language speakers. It is also an official language in Tanzania, Rwanda, Kenya, and Uganda and one of the three official languages of the East African Community (EAC). The language has 30 letters, including 24 Latin letters without characters (q and x) and six additional consonants (ch, dh, gh, ng’, sh, th) unique to Swahili pronunciation.

\subsection{Yoruba}
Over 40 million native speakers of Yoruba speak it in South-Western Nigeria and parts of Togo, Benin, and Ghana. It has 18 consonants, seven oral vowels, five nasal vowels, and syllabic nasals. The Yoruba language is tonal, with three tones: low, mid, and high. The tonal marks and underdots are referred to as diacritics and are needed for a word's correct pronunciation. The language's sentence structure is Subject-Verb-Object.

\section{Dataset}
We present two datasets: Inkuba-Mono (for monolingual pre-training) and Inkuba-Instruct (for instruction fine-tuning) for five languages listed in Section \ref{langs}.

\subsection{Inkuba-Mono Dataset} 
Inkuba-Mono\footnote{\url{https://huggingface.co/datasets/lelapa/Inkuba-Mono}} is a monolingual dataset collected from open-source repositories in the five languages to train the InkubaLM model. We collected open-source datasets for these five African languages from repositories on Hugging Face\footnote{\url{https://huggingface.co/datasets}}, Github\footnote{\url{https://github.com/}} and Zenodo\footnote{\url{https://zenodo.org/}}. After pre-processing and combining the raw dataset, we used 2.4 billion tokens to train the InkubaLM models.  Table \ref{mono_data} shows Inkuba-Mono dataset statistics.

\begin{table}[h!]
\centering
\resizebox{0.4\textwidth}{!}{
\begin{tabular}{l|r|r}
\hline
\textbf{Language} & \textbf{Number of sentences} & \textbf{Tokens}   \\ \hline
Hausa & 10.5 M & 345 M  \\ \hline
Yoruba & 2.2 M & 759.5 M   \\ \hline
Swahili &44.3 M& 1.2 B  \\ \hline
isiZulu & 9 M & 172.7 M \\ \hline
isiXhosa & 3 M & 62.5 M    \\ \hline \hline
\textbf{African only} & \textbf{69.3 M }&  \textbf{1.9 B }  \\ \hline
English & 18.2 M &  409.9 M   \\ \hline
French & 5.2 M &  174 M \\ \hline \hline 
\textbf{Total}  & \textbf{93.2 M}&  \textbf{ 2.4 B } \\ \hline
\end{tabular}
}
\caption{Dataset statistics for Inkuba-Mono dataset. M stands for Millions, and B stands for Billions. }
\label{mono_data}
\end{table}

\subsection{Inkuba-Instruct Dataset} \label{instr}
The Inkuba-Instruct\footnote{\url{https://huggingface.co/datasets/lelapa/Inkuba-instruct}} dataset is a comprehensive multilingual instruction dataset, combining several open-source downstream datasets designed to support a range of natural language processing tasks in these African languages. Our instruction dataset focused on six tasks for each of the five languages: Machine Translation, Sentiment Analysis, Named Entity Recognition (NER), Parts of Speech Tagging (POS), Question Answering, and Topic Classification.
Table \ref{dataset_sources} summarizes the datasets and their sources we used for each task, and Table \ref{dataset_sizes} summarizes the statistics of the instruction dataset of each language. It is important to note that we use English only for Machine Translation as a pivot language from and to the African languages.

\begin{table}[h!]
\centering
\begin{tabular}{l|r}
\hline
\textbf{Language} & \textbf{Number of samples} \\ \hline
Hausa & 5.8 M \\ \hline
Yoruba & 6.4 M \\ \hline
Swahili & 62.41 M \\ \hline
isiZulu & 16.20 M \\ \hline
isiXhosa & 25.35 M \\ \hline
English & 95.42 M \\ \hline
\end{tabular}
\caption{Dataset statistics for different languages. English is used as a pivot language for Machine Translation from ($eng \rightarrow xxx$) and to ($xxx \rightarrow eng$) African languages.}
\label{dataset_sizes}
\end{table}

\begin{table*}[h!]
\centering
\resizebox{\textwidth}{!}{
\begin{tabular}{|l|l|r|}
\hline
\textbf{Task} & \textbf{Datasets} & \textbf{Total Size (\# samples)} \\ \hline
Machine Translation & WMT-22-African \cite{nllbteam2022languageleftbehindscaling}, Mafand-MT \cite{adelani2022few}, Menyo-20k \cite{adelani2021effect} & 359 M \\ \hline
NER & MasakhaNER2 \cite{Adelani2022MasakhaNER2A}, Hausa VoA NER \cite{hedderich-etal-2020-transfer}, isiXhosa NER Corpus \cite{eiselen2016government} & 64k \\ \hline
POS & MasakhaPOS \cite{dione-etal-2023-masakhapos} & 6.5k \\ \hline
Question-Answering & AfriQA \cite{ogundepo2023afriqa} & 4.45k \\ \hline
Topic Classification & SIB-200 \cite{adelani2023sib200}, MasakhaNEWS \cite{Adelani2023MasakhaNEWS}, Hausa News Classification \cite{hedderich-etal-2020-transfer} & 22.8k \\ \hline
Sentiment Analysis & AfriSenti \cite{Muhammad2023AfriSentiAT,muhammad2023semeval}, NaijaSenti \cite{muhammad-etal-2022-naijasenti}, \href{https://huggingface.co/datasets/Davis/Swahili-tweet-sentiment}{Swahili-Tweet-Sentiment} & 46.62k \\ \hline
\end{tabular}
}
\caption{Sources of downstream datasets for the five African languages.}
\label{dataset_sources}
\end{table*}
\begin{figure*}[h!]
    \centering
    \includegraphics[width=\linewidth]{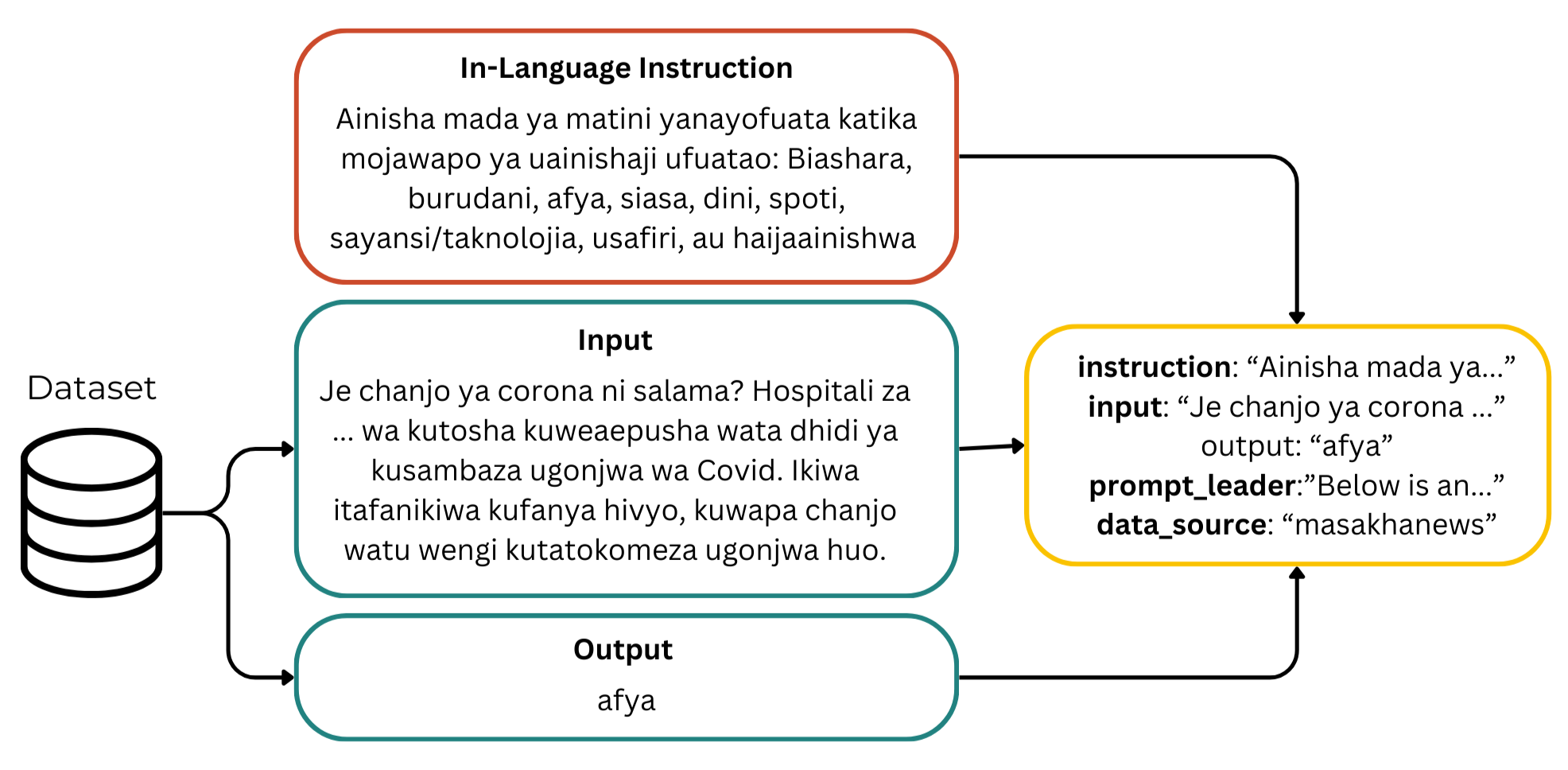}
    \caption{Example of the Swahili Topic Classification dataset converted into an Instruction dataset.}
    \label{fig:instrcut_dataset}
\end{figure*}

We created prompt templates for these tasks in English and manually translated them into the five African languages. We built the instruction datasets for the machine translation task in two directions ($xxx \rightarrow eng$ and $eng \rightarrow xxx$, where $xxx$ represents the African language). Regarding the Topic Classification and Sentiment Analysis tasks, we translated and used the labels in the respective target languages (i.e., if the label is `politics', then for Swahili, we use the Swahili translation of `politics'; if we switch to Hausa, we use the Hausa translation of `politics'). We did not perform this mapping for tasks such as NER and POS, as the labels are language agnostic. After generating the instruction inputs and targets for each task and language, we merged them and added a `task' column to make it easier to filter later. We split into `train', `dev' and `test' sets. Across all languages, merging tasks, we created a training instruction dataset of 148M samples, a validation set of 65M samples, and a testing set of size 55M samples. In Figure \ref{fig:instrcut_dataset}, we show an example of how we converted the Swahili Topic Classification dataset into an instruction dataset.

\section{InkubaLM} 
InkubaLM is the first decoder-only lightweight African language model with a 0.4B parameter model trained from scratch with an autoregressive language modeling objective for these five African languages. During training, we also included English and French datasets due to their prevalence in many African regions and the tendency for natural African languages to code-mix.

InkubaLM architecture and hyperparameters follow existing work \cite{thawakar2024mobillama}, with a slight modification by introducing multilingual capability and implementing custom Flash Attention \cite{dao2022flashattention} to enhance efficiency. This technique allows us to optimize the utilization of the compute resources, resulting in improved performance and efficiency during the training and inference process. During training, we incorporated Fully Sharded Data Parallel (FSDP) to efficiently utilize multi-GPU and multi-node setups. Table \ref{hyper} shows the InkubaLM-0.4B model architecture and hyperparameters used during training. 
\begin{table}[h!]
\centering
\begin{tabular}{l|r}
\hline
\textbf{Hyperparameter} & \textbf{Value} \\ \hline
Total Parameters & 0.422B \\ \hline
Hidden Size & 2048 \\ \hline
Intermediate Size (MLPs) & 5632 \\ \hline
Number of Attention Heads & 32 \\ \hline
Number of Hidden Layers & 8 \\ \hline
RMSNorm $\epsilon$ & $1 \times 10^{-5}$ \\ \hline
Max Seq Length & 2048 \\ \hline
Vocab Size & 61788 \\ \hline
\end{tabular}
\caption{InkubaLM-0.4B architecture and hyperparameters.}
\label{hyper}
\end{table}

\begin{figure*}[h!]
    \centering
    \includegraphics[width=\linewidth]{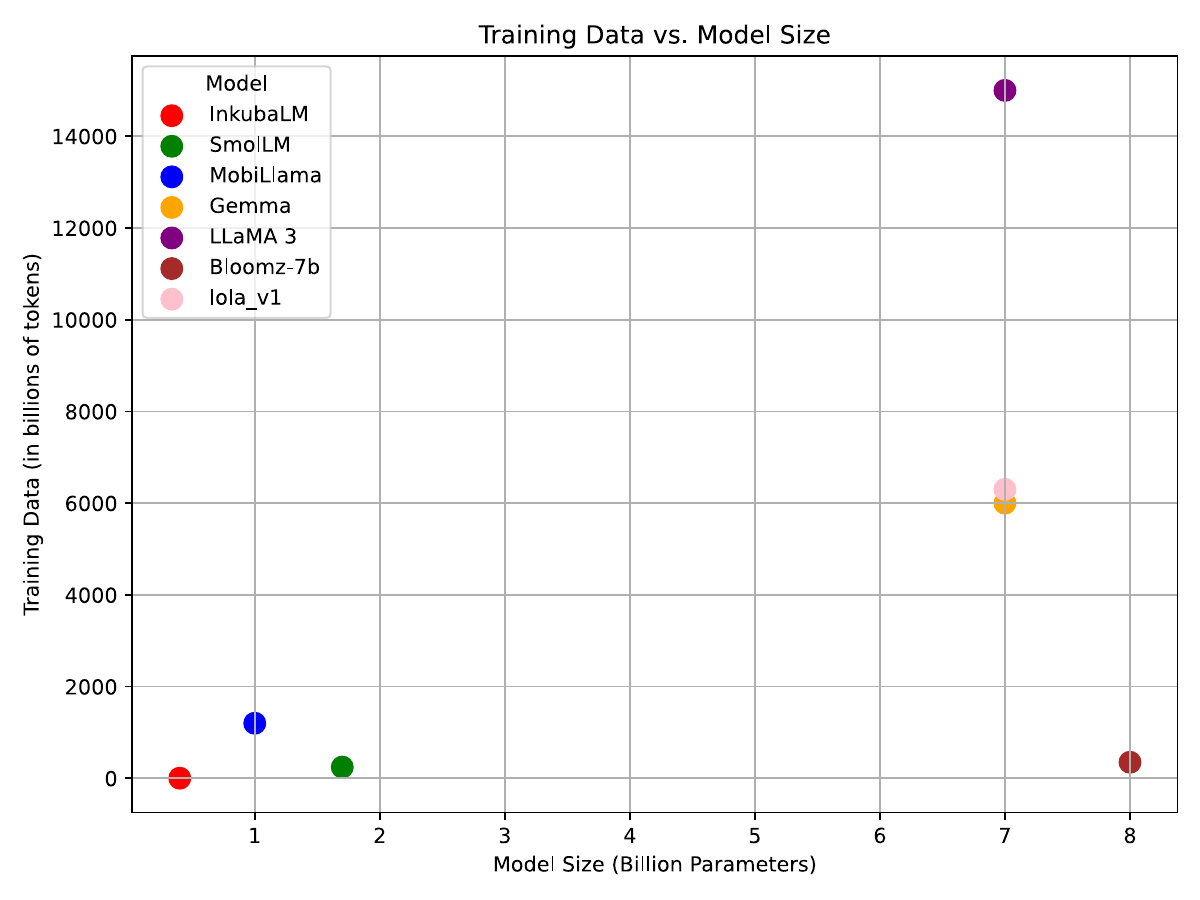}
    \caption{Training data(in billions of token) vs. Model size (in billions of parameters)}
    \label{fig:datavsmodel}
\end{figure*}

We employ Byte Pair Encoding (BPE) \cite{gage1994new} to train a multilingual tokenizer with a vocabulary size of 61788. The figure below shows different public models' training data and model sizes. In Figure \ref{fig:datavsmodel}, we compared the training dataset size used for models ranging from 0.4B up to 8B in terms of parameters. As shown in the Figure, our model is the smallest and has been trained using the least data from the compared models.

The environmental impact of training the InkubaLM-0.4B model was considered and measured using the machine learning impact calculator \footnote{\url{https://mlco2.github.io/impact\#compute}}. The training process on 8 A100 NVIDIA GPUs over 16 days resulted in an estimated carbon emission of 53.76 kg of CO2 equivalent. The training was performed on the Google Cloud Platform (GCP) in the Asia-Southeast1-C region, where the carbon footprint was carefully monitored to assess the environmental impact.

\section{Evaluation} 

\subsection{Evaluation Tool}

To evaluate models, we use the EleutherAI LM Evaluation Harness tool \cite{eval-harness} —- a popular evaluation framework that supports a wide range of zero- and few-shot evaluation tasks on autoregressive language models. In all evaluations, we use zero-shots and prompts in these languages; Section \ref{prompt} discusses our prompt designs.

\subsection{Model Selection}
We select open-source base models to evaluate their performance on tasks discussed in Section \ref{task}. We compare the performance of small, big, and multilingual models with our InkubaLM model. For small models, we use SmolLM \cite{allal2024SmolLM} with 1.7B parameters and MobiLlama \cite{thawakar2024mobillama} with 1B parameters, for Big models we use Gemma \cite{gemma_2024} with 7B parameters and LLaMa 3 \cite{dubey2024llama3herdmodels} with 8B parameters and for multilingual models, we use BLOOMZ with 7B parameters and lola\_v1\footnote{\url{https://huggingface.co/dice-research/lola_v1}} with 7.4B parameters.

\subsection{Tasks} \label{task}
We evaluate models with the Inkuba-Instruct (Section \ref{instr}) and IrokoBench \cite{adelani2024irokobench} datasets. From the Inkuba-Instruct dataset, we select sentiment analysis and machine translation tasks, as discussed in Section \ref{instr}; these tasks are created by combining task datasets from different open-source platforms. IrokoBench is a human-translated benchmark dataset that includes languages from various geographical regions of Africa. We use AfriXnli, a human-translated dataset for African languages from the English portion of XNLI \cite{conneau2018xnli}, and AfriMMLU, a human-translated dataset for the African language from MMLU \cite{hendrycks2020measuring}. 

\subsection{Prompts} \label{prompt}
For the Inkuba-Instruct dataset, we explore three prompts: \textbf{(1) Multiple Prompts (direct)}: the model is prompted using four slightly different prompts at random curated for each task in five languages.
\textbf{(2) Single prompt -- English (English)}: the model is prompted using one English prompt curated for each task 
\textbf{(3) Single prompt -- Native (native)}: the model is prompted using one prompt in the language curated for each task. Table \ref{tab:prompt} shows different prompt samples we use for initial evaluation.
\begin{table*}[h!]
    \centering
    \begin{tabular}{c|c|c|c}
         Tasks&  Prompts & langs &prompt\\ \hline
         Sentiment&Tafadhali tambua mawazo yaliyoonyeshwa kwenye   &\\
         & matini haya kwa  kutegemea miongozo ifuatayo:   &\\ &Chanya: ---, Hasi: ---, Wastani: --- $\{inputs\}$ & \\
         &Output: &swa&Native \\ \hline
         MT (swa-eng)&Tafsiri zifuatazo kutoka kwa Swahili hadi English. $\{inputs\}$& \\
         &Output: &swa&Native \\ \hline
         MT (eng-swa)&Tafsiri zifuatazo kutoka kwa English hadi Swahili. $\{inputs\}$ & \\
         &Output: &swa&Native \\ \hline
         Sentiment&Please identify the sentiment reflected in this   &\\
         & text based on the following guidelines:  &\\ &Positive:  ---, Negative: ---, Neutral: --- $\{inputs\}$ & \\
         &Output: &swa&English \\ \hline
          MT (eng-swa)&Translate the following from Swahili into English. $\{inputs\}$ & \\
         &Output: &swa&English \\ \hline
    \end{tabular}
    \caption{Sample prompt templates used for machine translation and sentiment analysis tasks}
    \label{tab:prompt}
\end{table*}
We follow the same template for the IrokoBench dataset used in their work. We report results for all tasks using \textbf{native} prompts.

\section{Results} \label{res}

\subsection{Sentiment Analysis}
The sentiment analysis results, presented in Table \ref{sentiment_results}, highlight the performance of various language models, including InkubaLM-0.4B, on three African languages: Swahili (swa), Hausa (hau), and Yoruba (yor). These results were obtained in a zero-shot setting using English prompts.

\begin{table}[h!]
\footnotesize
\begin{tabular}{lrrr|c}
\hline
\textbf{Model}   & \textbf{swa} & \textbf{hau} & \textbf{yor} & \textbf{AVG} \\ \hline
\multicolumn{5}{l}{\textit{Prompt LLMs in English Language}} \\ 
\textbf{InkubaLM-0.4B}  & \textbf{42.47}      & 22.25 &28.08  &30.93\\  \hline  \hline
SmolLM-1.7B   & 26.09     &31.97  &28.36  &28.80    \\ 
MobiLlama-1B & 37.2     &34.53  &32.89  &  34.87  \\ 
Gemma-7B & 14.42     & 36.16 & 26.17 &25.58  \\ 
LLaMa 3-8B& 19.48     &32.44  & 29.77 &27.23     \\ 
BLOOMZ-7B & 17.26     & 33.81 & 32.99 & 28.02    \\ 
lola\_v1-7.4B &  14.4    & 26.71 &28.16  & 22.42   \\ \hline
\end{tabular}
\caption{Sentiment Analysis results using prompt in English. The above results are zero-shot results only }
\label{sentiment_results}
\end{table}
\subsubsection{Key Observations}
\begin{itemize}
    \item \textbf{InkubaLM}: achieves an average score of \textbf{30.93}, and is particularly impressive in Swahili, where it scores \textbf{42.47}. This score is the highest among all the models tested, indicating that InkubaLM is particularly well-tuned for Sentiment Analysis in Swahili.
    \item \textbf{Comparison with Other Models}:
    \begin{itemize}
        \item \textbf{SmolLM-1.7B}: achieves a lower average score of \textbf{28.80}. This suggests that InkubaLM-0.4B's optimizations may be more effective for certain African languages, particularly Swahili.
        \item \textbf{MobiLlama-1B}: This model outperforms InkubaLM-0.4B on average, with a score of \textbf{34.87}, showing stronger performance in Hausa and Yoruba. However, it falls short in Swahili compared to InkubaLM-0.4B.
        \item \textbf{Gemma-7B} and \textbf{LLaMa 3-8B}: These models show mixed performance, with Gemma-7B excelling in Hausa but underperforming in Swahili. LLaMa 3-8B shows consistent, moderate performance across all languages.
        \item \textbf{BLOOMZ-7B} and \textbf{lola\_v1-7.4B}: These models generally perform worse, particularly in Swahili, where their scores are significantly lower than those of InkubaLM-0.4B.
    \end{itemize}
\end{itemize}

\subsubsection{Analysis}
The results indicate that InkubaLM is highly competitive in Sentiment Analysis, especially in Swahili. Its performance in this task demonstrates the effectiveness of the model's architecture and training optimizations, such as using Flash Attention and including multilingual capabilities. Despite its smaller size, InkubaLM outperforms or matches larger models' performance in key areas, making it a strong contender for tasks involving African languages. 
% Section \ref{sentcom} shows the average F1 score of the models and their parameter size for the sentiment analysis task.

\subsection{Machine Translation}
The machine translation results are split into two tasks: translating from English to African languages (Table \ref{eng_to_afr_mmt}) and from African languages to English (Table \ref{afr_to_eng_mmt}). The performance is measured using the BLEU score, a widely used metric for evaluating machine translation quality.

\begin{table*}[h!]
\footnotesize
\centering
\begin{tabular}{lrrrrr|c}
\hline
\textbf{Model}   & \textbf{swa} & \textbf{hau} & \textbf{yor} & \textbf{xho} & \textbf{zul} & \textbf{AVG} \\ \hline
\multicolumn{5}{l}{\textit{Prompt LLMs in African Language}} \\ 
\textbf{InkubaLM-0.4B}  &  3.44   & 4.67   &3.49     & 7.0    & \textbf{21}  &   8.15 \\  \hline \hline
SmolLM-1.7B    & 13.08    & 3.24    &  7.36   &  8.23  & 2.21  &  \underline{\underline {6.47}}  \\ 
MobiLlama-1B & 1.78    &  3.24   & 7.36    & 8.23    & 2.21 &  6.47  \\ 
Gemma-7B &  3.46   & 2    & 3.75    &  2.95   &  1.99  & 2.80 \\ 
LLaMa 3-8B&  3.17   & 4.17    &  5.11   &  5.45   & 2.83 & 3.82   \\ 
BLOOMZ-7B  & 3.28    &   1.73  & 3.85    & 3   &2.76   &  2.90  \\ 
lola\_v1-7.4B  & 5.63    & 14.5    &   9.3  &  8.87   & 9.7  &9.78   \\ \hline
\end{tabular}
\caption{Zero-shot BLEU score for English to African translation results using prompts in the African language.}
\label{eng_to_afr_mmt}
\end{table*}

\begin{table*}[h!]
\footnotesize
\centering
\begin{tabular}{lrrrrr|c}
\hline
\textbf{Model}   & \textbf{swa} & \textbf{hau} & \textbf{yor} & \textbf{xho} & \textbf{zul} & \textbf{AVG} \\ \hline
\multicolumn{5}{l}{\textit{Prompt LLMs in African Language}} \\ 
\textbf{InkubaLM-0.4B}  &  5.29   &   4.82  &  5.71   &  10.06   & 8.06 &  6.78 \\  \hline  \hline
SmolLM-1.7B    & 0.72    & 6.24    &  16.97   &  0.73   & 2.64 &  5.46  \\ 
MobiLlama-1B &  1.16   & 3.42   &  7.35    &  2.96   &4.72  &  3.92  \\ 
Gemma-7B & 1.24    & 0.81    & 5.58    & 0.76    &  0.11  &  1.790\\ 
LLaMa 3-8B&17.38     & 5.54    & 20.24    &  26.26   &15.89  &  17.06  \\ 
BLOOMZ-7B  &  0.99   &  11.26   & 5.21    &  3.75   & 15.91 & 7.42   \\ 
lola\_v1-7.4B  & 7.89    & 8.73    & 2.20    &  19.03   & 14.47  &  10.46 \\ \hline
\end{tabular}
\caption{Zero-shot BLEU score for African to English translation results using prompts in African language.}
\label{afr_to_eng_mmt}
\end{table*}

\subsubsection{English to African Languages (Zero-shot)}

\begin{itemize}
    \item \textbf{InkubaLM} achieves an average BLEU score of \textbf{8.15}, with its highest performance in isiZulu (\textbf{21}). The model shows moderate performance in the other languages, with BLEU \cite{papineni2002bleu} scores ranging from \textbf{3.44} in Swahili to \textbf{7.0} in Xhosa.
    \item \textbf{Comparison with other models}:
    \begin{itemize}
        \item \textbf{SmolLM-1.7B} and \textbf{MobiLlama-1B}: Both models achieve an average BLEU score of \textbf{6.47}. While SmolLM-1.7B performs better in Swahili, it significantly underperforms in isiZulu compared to InkubaLM-0.4B.
        \item \textbf{Gemma-7B} and \textbf{LLaMa 3-8B}: These models generally underperform in comparison to InkubaLM-0.4B, especially in isiZulu, where they achieve much lower BLEU scores.
        \item \textbf{lola\_v1-7.4B}: Despite having the highest average BLEU score (\textbf{9.78}), lola\_v1-7.4B shows inconsistent performance across different languages, with particularly high scores in isiZulu but lower scores in other languages.
    \end{itemize}
\end{itemize}

\subsubsection{Analysis}
InkubaLM-0.4B demonstrates strong performance in the English to African language translation task, particularly excelling in isiZulu. Its competitive average BLEU score suggests that the model is well-suited for specific languages, even when compared to larger models. However, its performance across different languages is variable, indicating that further refinement could enhance its overall translation capabilities while the model is optimized for particular tasks and languages.

\subsubsection{African Languages to English (Zero-shot)}

\begin{itemize}
    \item \textbf{InkubaLM-0.4B} shows moderate performance with an average BLEU score of \textbf{6.78}, with notable strengths in Xhosa (\textbf{10.06}) and Swahili (\textbf{5.29}).
    \item \textbf{Comparison with other models}:
    \begin{itemize}
        \item \textbf{LLaMa 3-8B} stands out with a significantly higher average BLEU score of \textbf{17.06}, particularly excelling in Xhosa and Swahili. This model outperforms InkubaLM-0.4B across all languages.
        \item \textbf{BLOOMZ-7B} and \textbf{lola\_v1-7.4B} also perform well, with lola\_v1-7.4B achieving the second-highest average score (\textbf{10.46}).
        \item \textbf{SmolLM-1.7B} and \textbf{MobiLlama-1B}: These models show mixed results, with relatively lower scores in several languages compared to InkubaLM-0.4B.
    \end{itemize}
\end{itemize}

\subsubsection{Analysis}
The results for African languages to English translation indicate that InkubaLM-0.4B performs reasonably well but is outperformed by larger models such as LLaMa 3-8B and lola\_v1-7.4B. The variability in scores suggests that while InkubaLM-0.4B is effective in some instances, particularly in translating Xhosa and Swahili, it may benefit from further tuning and optimization to improve its overall translation performance in this direction.

\subsection{AfriMMLU}

Table \ref{afrimmlu} presents the F1 scores for various models on the AfriMMLU task, using prompts in the five African languages. The models evaluated include InkubaLM-0.4B, SmolLM-1.7B, MobiLlama-1B, Gemma-7B, LLaMa 3-8B, BLOOMZ-7B, and lola\_v1-7.4B.

\subsubsection{Key Observations}
\begin{itemize}
    \item \textbf{InkubaLM-0.4B}: This model achieves an average F1 score of \textbf{26.16} across the five languages, with its best performance in Hausa (29.4) and Xhosa (27.4). It performs consistently across all languages, showing balanced capabilities.
    \item \textbf{Comparison with other models}:
    \begin{itemize}
        \item \textbf{Gemma-7B}: This model outperforms all others, with the highest average F1 score of \textbf{30.28}. It shows particularly strong performance in Swahili (34) and Zulu (32.4), indicating its effectiveness across various African languages.
        \item \textbf{LLaMa 3-8B}: With an average F1 score of \textbf{28.12}, this model also performs well, closely following Gemma-7B, and is particularly strong in Swahili (30.4) and Hausa (29.6).
        \item \textbf{BLOOMZ-7B}: This model achieves an average F1 score of \textbf{24.2}, with consistent but lower scores across all languages compared to InkubaLM-0.4B and the top-performing models.
        \item \textbf{SmolLM-1.7B} and \textbf{MobiLlama-1B}: These models have lower average F1 scores of \textbf{21.88} and \textbf{22.28}, respectively. They show weaker performance across most languages, particularly in Xhosa and Yoruba.
        \item \textbf{lola\_v1-7.4B}: Despite being a larger model, lola\_v1-7.4B achieves a relatively low average F1 score of \textbf{23.0}, with inconsistent performance, particularly lower in Yoruba (18.8) and Zulu (21).
    \end{itemize}
\end{itemize}

\subsubsection{Analysis}
The results indicate that InkubaLM-0.4B performs competitively on the AfriMMLU task, with an average F1 score that positions it in the mid-range among the models tested. While it does not outperform the better models such as Gemma-7B or LLaMa 3-8B, InkubaLM-0.4B shows consistent performance across all languages, making it a reliable option for tasks involving African languages.

The top-performing model, Gemma-7B, demonstrates the highest average F1 score, suggesting that larger model sizes, combined with practical multilingual training, can significantly enhance performance on complex tasks like AfriMMLU. However, the performance of InkubaLM-0.4B, despite its smaller size, highlights the effectiveness of its architecture and training optimizations, such as Flash Attention and multilingual capabilities.

Given these results, while larger models dominate raw performance, InkubaLM-0.4B offers a balanced trade-off between model size and effectiveness, particularly in resource-constrained settings where computational efficiency is a priority and data resources are limited. 
% Section \ref{afrimmlucom} shows the average F1 score of the models and their parameter size for the sentiment analysis task.

\begin{table*}[!t]
\footnotesize
\centering
\begin{tabular}{lrrrrr|c}
\hline
\textbf{Model}   & \textbf{swa} & \textbf{hau} & \textbf{yor} & \textbf{xho} & \textbf{zul} & \textbf{AVG} \\ \hline
\multicolumn{5}{l}{\textit{Prompt LLMs in African Language}} \\ 
\textbf{InkubaLM-0.4B}  & 25    & 29.4    & 24.8    & 27.4    & 24.2 & 26.16 \\ \hline \hline
SmolLM-1.7B    & 20.2    & 21.8    & 23.4    &19.2     & 24.8 &21.88    \\ 
MobiLlama-1B &19.8     & 20.2    & 24.2    &  25.2   &22  &   22.28 \\ 
Gemma-7B & 34    &  29   &  29   &  27   & 32.4   &30.28  \\ 
LLaMa 3-8B&  30.4   &29.6     &  28   &  24.6   & 28 &28.12    \\ 
BLOOMZ-7B  &26     & 22.2    & 25.6    &   23.2  & 24 & 24.2   \\ 
lola\_v1-7.4B  & 32.44    & 21.2    & 18.8    &21.6     & 21  & 23.0  \\ \hline
\end{tabular}
\caption{ F1 score for AfriMMLU results using prompt in African language. }
\label{afrimmlu}
\end{table*}

\subsection{AfriXnli}
Table \ref{afrixnli} presents the F1 scores for various models on the AfriXnli task, using prompts in the five African languages. The models evaluated include InkubaLM-0.4B, SmolLM-1.7B, MobiLlama-1B, Gemma-7B, LLaMa 3-8B, BLOOMZ-7B, and lola\_v1-7.4B.

\subsubsection{Key Observations}
\begin{itemize}
    \item \textbf{InkubaLM-0.4B}: InkubaLM-0.4B achieves an average F1 score of \textbf{33.47}, with consistent performance across all languages, ranging from 32.8 in Xhosa to 35.3 in Hausa. This consistency indicates the model's balanced capability across the different African languages.
    \item \textbf{Comparison with other models}:
    \begin{itemize}
        \item \textbf{SmolLM-1.7B}: This model achieves an average F1 score of \textbf{33.09}, performing similarly to InkubaLM-0.4B but slightly lower on average.
        \item \textbf{MobiLlama-1B}: With an average F1 score of \textbf{33.66}, MobiLlama-1B slightly outperforms InkubaLM-0.4B, showing strong performance across all languages with scores consistently above 32.5.
        \item \textbf{Gemma-7B}: This model achieves a significantly higher average F1 score of \textbf{37.5}, with particularly strong performance in Swahili (39.3) and Hausa (38.8), indicating its robustness across these languages.
        \item \textbf{LLaMa 3-8B}: With an average F1 score of \textbf{37.7}, LLaMa 3-8B also performs well, closely following Gemma-7B, and shows strong results across all languages.
        \item \textbf{BLOOMZ-7B}: This model stands out with the highest average F1 score of \textbf{41.52}, demonstrating exceptional performance in Swahili (45.8) and Yoruba (45.2).
        \item \textbf{lola\_v1-7.4B}: This model achieves an average F1 score of \textbf{34.13}, performing comparably to InkubaLM-0.4B, with relatively consistent results across all languages.
    \end{itemize}
\end{itemize}

\begin{table*}[h!]
\footnotesize
\centering
\begin{tabular}{lrrrrr|c}
\hline
\textbf{Model}   & \textbf{swa} & \textbf{hau} & \textbf{yor} & \textbf{xho} & \textbf{zul} & \textbf{AVG} \\ \hline
\multicolumn{5}{l}{\textit{Prompt LLMs in African Language}} \\ 
\textbf{InkubaLM-0.4B}  & 33    & 35.3    & 33.16    & 32.8 &33.1     & 33.47  \\ \hline \hline
SmolLM-1.7B    &  33.33 &   33.66  &  32.33   & 33.33   & 32.83  &   33.09 \\ 
MobiLlama-1B & 34.66  &  33.66   &  32.5   &  33.5   & 34 & 33.66   \\ 
Gemma-7B &  39.3 &  38.8   & 36.5    &    37.2   & 35.7   &37.5  \\ 
LLaMa 3-8B&  39.5 &  36   & 38.1    &    38.1  & 36.8 &  37.7  \\ 
BLOOMZ-7B  &  45.8& 36.5    &  45.2   &   39.8   &  40.3&  41.52  \\ 
lola\_v1-7.4B  & 36.83&33.5     & 33.16    &    33.83  &  33.33 &34.13   \\ \hline
\end{tabular}

\caption{AfriXnli results using prompt in African language.}
\label{afrixnli}
\end{table*}
\subsubsection{Analysis}
The results indicate that InkubaLM-0.4B is a strong performer in the AfriXnli task, achieving a consistent F1 score across all evaluated African languages. While it does not achieve the highest average score, its performance is close to that of other similarly sized models like SmolLM-1.7B and MobiLlama-1B.

The top-performing models, Gemma-7B, LLaMa 3-8B, and BLOOMZ-7B, demonstrate the benefits of larger model sizes. BLOOMZ-7B outperforms all others with an impressive average F1 score of 41.52. This suggests that while InkubaLM-0.4B is effective, larger models can offer significant performance advantages, especially in tasks like AfriXnli that require nuanced language understanding.

Overall, while InkubaLM-0.4B may not match the performance of the largest models, it offers a good balance of size and performance, making it a competitive option for African language processing tasks within resource-constrained environments. 
% Section \ref{afrixnlicom} shows the average F1 score of the models and their parameter size for the sentiment analysis task.

\section{Conclusion}
InkubaLM-0.4B has proven a reliable and efficient model tailored for African language processing, delivering robust performance across various tasks. Despite being a smaller model with only 0.4 billion parameters, it consistently holds its own against much larger models in tasks like Sentiment Analysis, Machine Translation, AfriMMLU, and AfriXnli. Its balanced performance across five key African languages (Swahili, Hausa, Yoruba, Xhosa, and Zulu) demonstrates the model's robustness and versatility, particularly in low-resource settings where such capabilities are critical.

In Sentiment Analysis, InkubaLM-0.4B stands out with its top performance in Swahili, outperforming all other models in this language. Similarly, in the Machine Translation task, the model shows strong results in isiZulu, further underscoring its ability to handle diverse linguistic contexts. The model’s design, which incorporates Flash Attention and supports multilingual capabilities, plays a significant role in its efficiency, allowing it to achieve competitive results without requiring extensive computational resources. This efficiency is critical in resource-constrained environments, where maximizing performance while minimizing computational costs is crucial.

While InkubaLM-0.4B may not always match the peak performance of the largest models like BLOOMZ-7B or LLaMa 3-8B, its consistent delivery of reliable results across all evaluated languages makes it a valuable tool for a wide range of applications. Despite its smaller size, the model’s ability to maintain competitive performance across various tasks highlights its potential as a go-to solution for African language processing, particularly in environments where resources are limited, and efficiency is paramount.

The imagination of NLP has traditionally been constrained by an emphasis on written histories and extensive text-based corpora, typically controlled by a few highly resourced institutions. SLMs like InkubaLM represent a significant step forward in exploring new possibilities beyond these conventional assumptions. In Africa, developing models we can train within the limitations of available computational resources and data is crucial. Such empowerment is essential for enabling local communities to address context-specific challenges, including broadened access to digital products and necessary services that would otherwise be out of reach. 

\section{Limitations}
The InkubaLM model has been trained on multilingual datasets but does have some limitations. It can understand and generate content in five African languages: Swahili, Yoruba, Hausa, isiZulu, and isiXhosa, as well as English and French. While it can generate text on various topics, the resulting content may not always be entirely accurate, logically consistent, or free from biases found in the training data. Additionally, the model may sometimes use different languages when generating text. Nonetheless, this model is intended to be a foundational tool to aid research in African languages.
\section{Ethical Considerations and Risks}
InkubaLM is a small LM developed for five African languages. The model is evaluated only in sentiment analysis, machine translation, AfriMMLU, and AfriXNLI tasks and has yet to cover all possible evaluation scenarios. Similar to other language models, it is impossible to predict all of InkubaLM's potential outputs in advance, and in some cases, the model may produce inaccurate, biased, or objectionable responses. Therefore, before using the model in any application, the users should conduct safety testing and tuning tailored to their intended use.
\section*{Acknowledgments}
We thank Microsoft AI4Good lab\footnote{\url{https://www.microsoft.com/en-us/research/group/ai-for-good-research-lab/}} for the compute credits to train the above model. This work would not have been possible without your sponsorship.

\bibliography{custom}

\begin{thebibliography}{45}
\providecommand{\natexlab}[1]{#1}

\bibitem[{Adelani et~al.(2021)Adelani, Ruiter, Alabi, Adebonojo, Ayeni, Adeyemi, Awokoya, and Espa{\~n}a-Bonet}]{adelani2021effect}
David~I Adelani, Dana Ruiter, Jesujoba~O Alabi, Damilola Adebonojo, Adesina Ayeni, Mofe Adeyemi, Ayodele Awokoya, and Cristina Espa{\~n}a-Bonet. 2021.
\newblock The effect of domain and diacritics in yor$\backslash$ub$\backslash$'a-english neural machine translation.
\newblock \emph{arXiv preprint arXiv:2103.08647}.

\bibitem[{Adelani et~al.(2022{\natexlab{a}})Adelani, Alabi, Fan, Kreutzer, Shen, Reid, Ruiter, Klakow, Nabende, Chang et~al.}]{adelani2022few}
David~Ifeoluwa Adelani, Jesujoba~Oluwadara Alabi, Angela Fan, Julia Kreutzer, Xiaoyu Shen, Machel Reid, Dana Ruiter, Dietrich Klakow, Peter Nabende, Ernie Chang, et~al. 2022{\natexlab{a}}.
\newblock A few thousand translations go a long way! leveraging pre-trained models for african news translation.
\newblock \emph{arXiv preprint arXiv:2205.02022}.

\bibitem[{Adelani et~al.(2023{\natexlab{a}})Adelani, Liu, Shen, Vassilyev, Alabi, Mao, Gao, and Lee}]{adelani2023sib200}
David~Ifeoluwa Adelani, Hannah Liu, Xiaoyu Shen, Nikita Vassilyev, Jesujoba~O. Alabi, Yanke Mao, Haonan Gao, and Annie En-Shiun Lee. 2023{\natexlab{a}}.
\newblock \href {https://arxiv.org/abs/2309.07445} {Sib-200: A simple, inclusive, and big evaluation dataset for topic classification in 200+ languages and dialects}.
\newblock \emph{Preprint}, arXiv:2309.07445.

\bibitem[{Adelani et~al.(2023{\natexlab{b}})Adelani, Masiak, Azime, Alabi, Tonja, Mwase, Ogundepo, Dossou, Oladipo, Nixdorf, Emezue, al~azzawi, Sibanda, David, Ndolela, Mukiibi, Ajayi, Ngoli, Odhiambo, Owodunni, Obiefuna, Muhammad, Abdullahi, Yigezu, Gwadabe, Abdulmumin, Bame, Awoyomi, Shode, Adelani, Kailani, Omotayo, Adeeko, Abeeb, Aremu, Samuel, Siro, Kimotho, Ogbu, Mbonu, Chukwuneke, Fanijo, Ojo, Awosan, Guge, Sari, Nyatsine, Sidume, Yousuf, Oduwole, Kimanuka, Tshinu, Diko, Nxakama, Johar, Gebre, Mohamed, Mohamed, Hassan, Mehamed, Ngabire, , and Stenetorp}]{Adelani2023MasakhaNEWS}
David~Ifeoluwa Adelani, Marek Masiak, Israel~Abebe Azime, Jesujoba~Oluwadara Alabi, Atnafu~Lambebo Tonja, Christine Mwase, Odunayo Ogundepo, Bonaventure F.~P. Dossou, Akintunde Oladipo, Doreen Nixdorf, Chris~Chinenye Emezue, Sana~Sabah al~azzawi, Blessing~K. Sibanda, Davis David, Lolwethu Ndolela, Jonathan Mukiibi, Tunde~Oluwaseyi Ajayi, Tatiana~Moteu Ngoli, Brian Odhiambo, Abraham~Toluwase Owodunni, Nnaemeka~C. Obiefuna, Shamsuddeen~Hassan Muhammad, Saheed~Salahudeen Abdullahi, Mesay~Gemeda Yigezu, Tajuddeen Gwadabe, Idris Abdulmumin, Mahlet~Taye Bame, Oluwabusayo~Olufunke Awoyomi, Iyanuoluwa Shode, Tolulope~Anu Adelani, Habiba~Abdulganiy Kailani, Abdul-Hakeem Omotayo, Adetola Adeeko, Afolabi Abeeb, Anuoluwapo Aremu, Olanrewaju Samuel, Clemencia Siro, Wangari Kimotho, Onyekachi~Raphael Ogbu, Chinedu~E. Mbonu, Chiamaka~I. Chukwuneke, Samuel Fanijo, Jessica Ojo, Oyinkansola~F. Awosan, Tadesse~Kebede Guge, Sakayo~Toadoum Sari, Pamela Nyatsine, Freedmore Sidume, Oreen Yousuf, Mardiyyah Oduwole, Ussen Kimanuka,
  Kanda~Patrick Tshinu, Thina Diko, Siyanda Nxakama, Abdulmejid~Tuni Johar, Sinodos Gebre, Muhidin Mohamed, Shafie~Abdi Mohamed, Fuad~Mire Hassan, Moges~Ahmed Mehamed, Evrard Ngabire, , and Pontus Stenetorp. 2023{\natexlab{b}}.
\newblock Masakhanews: News topic classification for african languages.
\newblock \emph{ArXiv}.

\bibitem[{Adelani et~al.(2022{\natexlab{b}})Adelani, Neubig, Ruder, Rijhwani, Beukman, Palen-Michel, Lignos, Alabi, Muhammad, Nabende, Dione, Bukula, Mabuya, Dossou, Sibanda, Buzaaba, Mukiibi, Kalipe, Mbaye, Taylor, Kabore, Emezue, Aremu, Ogayo, Gitau, Munkoh-Buabeng, Koagne, Tapo, Macucwa, Marivate, Mboning, Gwadabe, Adewumi, Ahia, Nakatumba-Nabende, Mokono, Ezeani, Chukwuneke, Adeyemi, Hacheme, Abdulmumin, Ogundepo, Yousuf, Ngoli, and Klakow}]{Adelani2022MasakhaNER2A}
David~Ifeoluwa Adelani, Graham Neubig, Sebastian Ruder, Shruti Rijhwani, Michael Beukman, Chester Palen-Michel, Constantine Lignos, Jesujoba~Oluwadara Alabi, Shamsuddeen~Hassan Muhammad, Peter Nabende, Cheikh M.~Bamba Dione, Andiswa Bukula, Rooweither Mabuya, Bonaventure F.~P. Dossou, Blessing~K. Sibanda, Happy Buzaaba, Jonathan Mukiibi, Godson Kalipe, Derguene Mbaye, Amelia Taylor, Fatoumata Kabore, Chris~C. Emezue, Anuoluwapo Aremu, Perez Ogayo, Catherine~W. Gitau, Edwin Munkoh-Buabeng, Victoire~Memdjokam Koagne, Allahsera~Auguste Tapo, Tebogo Macucwa, Vukosi Marivate, Elvis Mboning, Tajuddeen~R. Gwadabe, Tosin~P. Adewumi, Orevaoghene Ahia, Joyce Nakatumba-Nabende, Neo~L. Mokono, Ignatius~M Ezeani, Chiamaka~Ijeoma Chukwuneke, Mofetoluwa Adeyemi, Gilles Hacheme, Idris Abdulmumin, Odunayo Ogundepo, Oreen Yousuf, Tatiana~Moteu Ngoli, and Dietrich Klakow. 2022{\natexlab{b}}.
\newblock Masakhaner 2.0: Africa-centric transfer learning for named entity recognition.
\newblock \emph{ArXiv}, abs/2210.12391.

\bibitem[{Adelani et~al.(2024{\natexlab{a}})Adelani, Ojo, Azime, Zhuang, Alabi, He, Ochieng, Hooker, Bukula, Lee, Chukwuneke, Buzaaba, Sibanda, Kalipe, Mukiibi, Kabongo, Yuehgoh, Setaka, Ndolela, Odu, Mabuya, Muhammad, Osei, Samb, Guge, and Stenetorp}]{adelani2024irokobenchnewbenchmarkafrican}
David~Ifeoluwa Adelani, Jessica Ojo, Israel~Abebe Azime, Jian~Yun Zhuang, Jesujoba~O. Alabi, Xuanli He, Millicent Ochieng, Sara Hooker, Andiswa Bukula, En-Shiun~Annie Lee, Chiamaka Chukwuneke, Happy Buzaaba, Blessing Sibanda, Godson Kalipe, Jonathan Mukiibi, Salomon Kabongo, Foutse Yuehgoh, Mmasibidi Setaka, Lolwethu Ndolela, Nkiruka Odu, Rooweither Mabuya, Shamsuddeen~Hassan Muhammad, Salomey Osei, Sokhar Samb, Tadesse~Kebede Guge, and Pontus Stenetorp. 2024{\natexlab{a}}.
\newblock \href {https://arxiv.org/abs/2406.03368} {Irokobench: A new benchmark for african languages in the age of large language models}.
\newblock \emph{Preprint}, arXiv:2406.03368.

\bibitem[{Adelani et~al.(2024{\natexlab{b}})Adelani, Ojo, Azime, Zhuang, Alabi, He, Ochieng, Hooker, Bukula, Lee et~al.}]{adelani2024irokobench}
David~Ifeoluwa Adelani, Jessica Ojo, Israel~Abebe Azime, Jian~Yun Zhuang, Jesujoba~O Alabi, Xuanli He, Millicent Ochieng, Sara Hooker, Andiswa Bukula, En-Shiun~Annie Lee, et~al. 2024{\natexlab{b}}.
\newblock Irokobench: A new benchmark for african languages in the age of large language models.
\newblock \emph{arXiv preprint arXiv:2406.03368}.

\bibitem[{Allal et~al.(2024)Allal, Lozhkov, Bakouch, von Werra, and Wolf}]{allal2024SmolLM}
Loubna~Ben Allal, Anton Lozhkov, Elie Bakouch, Leandro von Werra, and Thomas Wolf. 2024.
\newblock Smollm - blazingly fast and remarkably powerful.

\bibitem[{Azime et~al.(2024)Azime, Fuge, Tonja, Belay, Wassie, Jada, Chanie, Sewunetie, and Yimam}]{azime2024enhancing}
Israel~Abebe Azime, Mitiku~Yohannes Fuge, Atnafu~Lambebo Tonja, Tadesse~Destaw Belay, Aman~Kassahun Wassie, Eyasu~Shiferaw Jada, Yonas Chanie, Walelign~Tewabe Sewunetie, and Seid~Muhie Yimam. 2024.
\newblock \href {https://openreview.net/forum?id=Zem7T1KTJs} {Enhancing amharic-llama: Integrating task specific and generative datasets}.
\newblock In \emph{5th Workshop on African Natural Language Processing}.

\bibitem[{Brei et~al.(2024)Brei, Frey, and Meyer}]{brei2024leveraging}
Felix Brei, Johannes Frey, and Lars-Peter Meyer. 2024.
\newblock Leveraging small language models for text2sparql tasks to improve the resilience of ai assistance.
\newblock \emph{arXiv preprint arXiv:2405.17076}.

\bibitem[{Brown et~al.(2020)Brown, Mann, Ryder, Subbiah, Kaplan, Dhariwal, Neelakantan, Shyam, Sastry, Askell, Agarwal, Herbert{-}Voss, Krueger, Henighan, Child, Ramesh, Ziegler, Wu, Winter, Hesse, Chen, Sigler, Litwin, Gray, Chess, Clark, Berner, McCandlish, Radford, Sutskever, and Amodei}]{GPT}
Tom~B. Brown, Benjamin Mann, Nick Ryder, Melanie Subbiah, Jared Kaplan, Prafulla Dhariwal, Arvind Neelakantan, Pranav Shyam, Girish Sastry, Amanda Askell, Sandhini Agarwal, Ariel Herbert{-}Voss, Gretchen Krueger, Tom Henighan, Rewon Child, Aditya Ramesh, Daniel~M. Ziegler, Jeffrey Wu, Clemens Winter, Christopher Hesse, Mark Chen, Eric Sigler, Mateusz Litwin, Scott Gray, Benjamin Chess, Jack Clark, Christopher Berner, Sam McCandlish, Alec Radford, Ilya Sutskever, and Dario Amodei. 2020.
\newblock \href {https://arxiv.org/abs/2005.14165} {Language models are few-shot learners}.
\newblock \emph{CoRR}, abs/2005.14165.

\bibitem[{Conneau et~al.(2019)Conneau, Khandelwal, Goyal, Chaudhary, Wenzek, Guzm{\'{a}}n, Grave, Ott, Zettlemoyer, and Stoyanov}]{XLM-R}
Alexis Conneau, Kartikay Khandelwal, Naman Goyal, Vishrav Chaudhary, Guillaume Wenzek, Francisco Guzm{\'{a}}n, Edouard Grave, Myle Ott, Luke Zettlemoyer, and Veselin Stoyanov. 2019.
\newblock \href {https://arxiv.org/abs/1911.02116} {Unsupervised cross-lingual representation learning at scale}.
\newblock \emph{CoRR}, abs/1911.02116.

\bibitem[{Conneau et~al.(2018)Conneau, Lample, Rinott, Williams, Bowman, Schwenk, and Stoyanov}]{conneau2018xnli}
Alexis Conneau, Guillaume Lample, Ruty Rinott, Adina Williams, Samuel~R Bowman, Holger Schwenk, and Veselin Stoyanov. 2018.
\newblock Xnli: Evaluating cross-lingual sentence representations.
\newblock \emph{arXiv preprint arXiv:1809.05053}.

\bibitem[{Dao et~al.(2022)Dao, Fu, Ermon, Rudra, and R{\'e}}]{dao2022flashattention}
Tri Dao, Dan Fu, Stefano Ermon, Atri Rudra, and Christopher R{\'e}. 2022.
\newblock Flashattention: Fast and memory-efficient exact attention with io-awareness.
\newblock \emph{Advances in Neural Information Processing Systems}, 35:16344--16359.

\bibitem[{Devlin et~al.(2018)Devlin, Chang, Lee, and Toutanova}]{BERT}
Jacob Devlin, Ming{-}Wei Chang, Kenton Lee, and Kristina Toutanova. 2018.
\newblock \href {https://arxiv.org/abs/1810.04805} {{BERT:} pre-training of deep bidirectional transformers for language understanding}.
\newblock \emph{CoRR}, abs/1810.04805.

\bibitem[{Dione et~al.(2023)Dione, Adelani, Nabende, Alabi, Sindane, Buzaaba, Muhammad, Emezue, Ogayo, Aremu, Gitau, Mbaye, Mukiibi, Sibanda, Dossou, Bukula, Mabuya, Tapo, Munkoh-Buabeng, Memdjokam~Koagne, Ouoba~Kabore, Taylor, Kalipe, Macucwa, Marivate, Gwadabe, Elvis, Onyenwe, Atindogbe, Adelani, Akinade, Samuel, Nahimana, Musabeyezu, Niyomutabazi, Chimhenga, Gotosa, Mizha, Agbolo, Traore, Uchechukwu, Yusuf, Abdullahi, and Klakow}]{dione-etal-2023-masakhapos}
Cheikh M.~Bamba Dione, David~Ifeoluwa Adelani, Peter Nabende, Jesujoba Alabi, Thapelo Sindane, Happy Buzaaba, Shamsuddeen~Hassan Muhammad, Chris~Chinenye Emezue, Perez Ogayo, Anuoluwapo Aremu, Catherine Gitau, Derguene Mbaye, Jonathan Mukiibi, Blessing Sibanda, Bonaventure F.~P. Dossou, Andiswa Bukula, Rooweither Mabuya, Allahsera~Auguste Tapo, Edwin Munkoh-Buabeng, Victoire Memdjokam~Koagne, Fatoumata Ouoba~Kabore, Amelia Taylor, Godson Kalipe, Tebogo Macucwa, Vukosi Marivate, Tajuddeen Gwadabe, Mboning~Tchiaze Elvis, Ikechukwu Onyenwe, Gratien Atindogbe, Tolulope Adelani, Idris Akinade, Olanrewaju Samuel, Marien Nahimana, Th{'e}og{`e}ne Musabeyezu, Emile Niyomutabazi, Ester Chimhenga, Kudzai Gotosa, Patrick Mizha, Apelete Agbolo, Seydou Traore, Chinedu Uchechukwu, Aliyu Yusuf, Muhammad Abdullahi, and Dietrich Klakow. 2023.
\newblock \href {https://doi.org/10.18653/v1/2023.acl-long.609} {{M}asakha{POS}: Part-of-speech tagging for typologically diverse {A}frican languages}.
\newblock In \emph{Proceedings of the 61st Annual Meeting of the Association for Computational Linguistics (Volume 1: Long Papers)}, pages 10883--10900, Toronto, Canada. Association for Computational Linguistics.

\bibitem[{Dubey et~al.(2024)Dubey, Jauhri, Pandey, Kadian, Al-Dahle, Letman, Mathur, Schelten, Yang, Fan, Goyal, Hartshorn, Yang, Mitra, Sravankumar, Korenev, Hinsvark, Rao, Zhang, Rodriguez, Gregerson, Spataru, Roziere, Biron, Tang, Chern, Caucheteux, Nayak, Bi, Marra, McConnell, Keller, Touret, Wu, Wong, Ferrer, Nikolaidis, Allonsius, Song, Pintz, Livshits, Esiobu, Choudhary, Mahajan, Garcia-Olano, Perino, Hupkes, Lakomkin, AlBadawy, Lobanova, Dinan, Smith, Radenovic, Zhang, Synnaeve, Lee, Anderson, Nail, Mialon, Pang, Cucurell, Nguyen, Korevaar, Xu, Touvron, Zarov, Ibarra, Kloumann, Misra, Evtimov, Copet, Lee, Geffert, Vranes, Park, Mahadeokar, Shah, van~der Linde, Billock, Hong, Lee, Fu, Chi, Huang, Liu, Wang, Yu, Bitton, Spisak, Park, Rocca, Johnstun, Saxe, Jia, Alwala, Upasani, Plawiak, Li, Heafield, Stone, El-Arini, Iyer, Malik, Chiu, Bhalla, Rantala-Yeary, van~der Maaten, Chen, Tan, Jenkins, Martin, Madaan, Malo, Blecher, Landzaat, de~Oliveira, Muzzi, Pasupuleti, Singh, Paluri, Kardas, Oldham, Rita,
  Pavlova, Kambadur, Lewis, Si, Singh, Hassan, Goyal, Torabi, Bashlykov, Bogoychev, Chatterji, Duchenne, Çelebi, Alrassy, Zhang, Li, Vasic, Weng, Bhargava, Dubal, Krishnan, Koura, Xu, He, Dong, Srinivasan, Ganapathy, Calderer, Cabral, Stojnic, Raileanu, Girdhar, Patel, Sauvestre, Polidoro, Sumbaly, Taylor, Silva, Hou, Wang, Hosseini, Chennabasappa, Singh, Bell, Kim, Edunov, Nie, Narang, Raparthy, Shen, Wan, Bhosale, Zhang, Vandenhende, Batra, Whitman, Sootla, Collot, Gururangan, Borodinsky, Herman, Fowler, Sheasha, Georgiou, Scialom, Speckbacher, Mihaylov, Xiao, Karn, Goswami, Gupta, Ramanathan, Kerkez, Gonguet, Do, Vogeti, Petrovic, Chu, Xiong, Fu, Meers, Martinet, Wang, Tan, Xie, Jia, Wang, Goldschlag, Gaur, Babaei, Wen, Song, Zhang, Li, Mao, Coudert, Yan, Chen, Papakipos, Singh, Grattafiori, Jain, Kelsey, Shajnfeld, Gangidi, Victoria, Goldstand, Menon, Sharma, Boesenberg, Vaughan, Baevski, Feinstein, Kallet, Sangani, Yunus, Lupu, Alvarado, Caples, Gu, Ho, Poulton, Ryan, Ramchandani, Franco, Saraf,
  Chowdhury, Gabriel, Bharambe, Eisenman, Yazdan, James, Maurer, Leonhardi, Huang, Loyd, Paola, Paranjape, Liu, Wu, Ni, Hancock, Wasti, Spence, Stojkovic, Gamido, Montalvo, Parker, Burton, Mejia, Wang, Kim, Zhou, Hu, Chu, Cai, Tindal, Feichtenhofer, Civin, Beaty, Kreymer, Li, Wyatt, Adkins, Xu, Testuggine, David, Parikh, Liskovich, Foss, Wang, Le, Holland, Dowling, Jamil, Montgomery, Presani, Hahn, Wood, Brinkman, Arcaute, Dunbar, Smothers, Sun, Kreuk, Tian, Ozgenel, Caggioni, Guzmán, Kanayet, Seide, Florez, Schwarz, Badeer, Swee, Halpern, Thattai, Herman, Sizov, Guangyi, Zhang, Lakshminarayanan, Shojanazeri, Zou, Wang, Zha, Habeeb, Rudolph, Suk, Aspegren, Goldman, Damlaj, Molybog, Tufanov, Veliche, Gat, Weissman, Geboski, Kohli, Asher, Gaya, Marcus, Tang, Chan, Zhen, Reizenstein, Teboul, Zhong, Jin, Yang, Cummings, Carvill, Shepard, McPhie, Torres, Ginsburg, Wang, Wu, U, Saxena, Prasad, Khandelwal, Zand, Matosich, Veeraraghavan, Michelena, Li, Huang, Chawla, Lakhotia, Huang, Chen, Garg, A, Silva, Bell,
  Zhang, Guo, Yu, Moshkovich, Wehrstedt, Khabsa, Avalani, Bhatt, Tsimpoukelli, Mankus, Hasson, Lennie, Reso, Groshev, Naumov, Lathi, Keneally, Seltzer, Valko, Restrepo, Patel, Vyatskov, Samvelyan, Clark, Macey, Wang, Hermoso, Metanat, Rastegari, Bansal, Santhanam, Parks, White, Bawa, Singhal, Egebo, Usunier, Laptev, Dong, Zhang, Cheng, Chernoguz, Hart, Salpekar, Kalinli, Kent, Parekh, Saab, Balaji, Rittner, Bontrager, Roux, Dollar, Zvyagina, Ratanchandani, Yuvraj, Liang, Alao, Rodriguez, Ayub, Murthy, Nayani, Mitra, Li, Hogan, Battey, Wang, Maheswari, Howes, Rinott, Bondu, Datta, Chugh, Hunt, Dhillon, Sidorov, Pan, Verma, Yamamoto, Ramaswamy, Lindsay, Lindsay, Feng, Lin, Zha, Shankar, Zhang, Zhang, Wang, Agarwal, Sajuyigbe, Chintala, Max, Chen, Kehoe, Satterfield, Govindaprasad, Gupta, Cho, Virk, Subramanian, Choudhury, Goldman, Remez, Glaser, Best, Kohler, Robinson, Li, Zhang, Matthews, Chou, Shaked, Vontimitta, Ajayi, Montanez, Mohan, Kumar, Mangla, Albiero, Ionescu, Poenaru, Mihailescu, Ivanov, Li, Wang,
  Jiang, Bouaziz, Constable, Tang, Wang, Wu, Wang, Xia, Wu, Gao, Chen, Hu, Jia, Qi, Li, Zhang, Zhang, Adi, Nam, Yu, Wang, Hao, Qian, He, Rait, DeVito, Rosnbrick, Wen, Yang, and Zhao}]{dubey2024llama3herdmodels}
Abhimanyu Dubey, Abhinav Jauhri, Abhinav Pandey, Abhishek Kadian, Ahmad Al-Dahle, Aiesha Letman, Akhil Mathur, Alan Schelten, Amy Yang, Angela Fan, Anirudh Goyal, Anthony Hartshorn, Aobo Yang, Archi Mitra, Archie Sravankumar, Artem Korenev, Arthur Hinsvark, Arun Rao, Aston Zhang, Aurelien Rodriguez, Austen Gregerson, Ava Spataru, Baptiste Roziere, Bethany Biron, Binh Tang, Bobbie Chern, Charlotte Caucheteux, Chaya Nayak, Chloe Bi, Chris Marra, Chris McConnell, Christian Keller, Christophe Touret, Chunyang Wu, Corinne Wong, Cristian~Canton Ferrer, Cyrus Nikolaidis, Damien Allonsius, Daniel Song, Danielle Pintz, Danny Livshits, David Esiobu, Dhruv Choudhary, Dhruv Mahajan, Diego Garcia-Olano, Diego Perino, Dieuwke Hupkes, Egor Lakomkin, Ehab AlBadawy, Elina Lobanova, Emily Dinan, Eric~Michael Smith, Filip Radenovic, Frank Zhang, Gabriel Synnaeve, Gabrielle Lee, Georgia~Lewis Anderson, Graeme Nail, Gregoire Mialon, Guan Pang, Guillem Cucurell, Hailey Nguyen, Hannah Korevaar, Hu~Xu, Hugo Touvron, Iliyan Zarov,
  Imanol~Arrieta Ibarra, Isabel Kloumann, Ishan Misra, Ivan Evtimov, Jade Copet, Jaewon Lee, Jan Geffert, Jana Vranes, Jason Park, Jay Mahadeokar, Jeet Shah, Jelmer van~der Linde, Jennifer Billock, Jenny Hong, Jenya Lee, Jeremy Fu, Jianfeng Chi, Jianyu Huang, Jiawen Liu, Jie Wang, Jiecao Yu, Joanna Bitton, Joe Spisak, Jongsoo Park, Joseph Rocca, Joshua Johnstun, Joshua Saxe, Junteng Jia, Kalyan~Vasuden Alwala, Kartikeya Upasani, Kate Plawiak, Ke~Li, Kenneth Heafield, Kevin Stone, Khalid El-Arini, Krithika Iyer, Kshitiz Malik, Kuenley Chiu, Kunal Bhalla, Lauren Rantala-Yeary, Laurens van~der Maaten, Lawrence Chen, Liang Tan, Liz Jenkins, Louis Martin, Lovish Madaan, Lubo Malo, Lukas Blecher, Lukas Landzaat, Luke de~Oliveira, Madeline Muzzi, Mahesh Pasupuleti, Mannat Singh, Manohar Paluri, Marcin Kardas, Mathew Oldham, Mathieu Rita, Maya Pavlova, Melanie Kambadur, Mike Lewis, Min Si, Mitesh~Kumar Singh, Mona Hassan, Naman Goyal, Narjes Torabi, Nikolay Bashlykov, Nikolay Bogoychev, Niladri Chatterji, Olivier
  Duchenne, Onur Çelebi, Patrick Alrassy, Pengchuan Zhang, Pengwei Li, Petar Vasic, Peter Weng, Prajjwal Bhargava, Pratik Dubal, Praveen Krishnan, Punit~Singh Koura, Puxin Xu, Qing He, Qingxiao Dong, Ragavan Srinivasan, Raj Ganapathy, Ramon Calderer, Ricardo~Silveira Cabral, Robert Stojnic, Roberta Raileanu, Rohit Girdhar, Rohit Patel, Romain Sauvestre, Ronnie Polidoro, Roshan Sumbaly, Ross Taylor, Ruan Silva, Rui Hou, Rui Wang, Saghar Hosseini, Sahana Chennabasappa, Sanjay Singh, Sean Bell, Seohyun~Sonia Kim, Sergey Edunov, Shaoliang Nie, Sharan Narang, Sharath Raparthy, Sheng Shen, Shengye Wan, Shruti Bhosale, Shun Zhang, Simon Vandenhende, Soumya Batra, Spencer Whitman, Sten Sootla, Stephane Collot, Suchin Gururangan, Sydney Borodinsky, Tamar Herman, Tara Fowler, Tarek Sheasha, Thomas Georgiou, Thomas Scialom, Tobias Speckbacher, Todor Mihaylov, Tong Xiao, Ujjwal Karn, Vedanuj Goswami, Vibhor Gupta, Vignesh Ramanathan, Viktor Kerkez, Vincent Gonguet, Virginie Do, Vish Vogeti, Vladan Petrovic, Weiwei Chu,
  Wenhan Xiong, Wenyin Fu, Whitney Meers, Xavier Martinet, Xiaodong Wang, Xiaoqing~Ellen Tan, Xinfeng Xie, Xuchao Jia, Xuewei Wang, Yaelle Goldschlag, Yashesh Gaur, Yasmine Babaei, Yi~Wen, Yiwen Song, Yuchen Zhang, Yue Li, Yuning Mao, Zacharie~Delpierre Coudert, Zheng Yan, Zhengxing Chen, Zoe Papakipos, Aaditya Singh, Aaron Grattafiori, Abha Jain, Adam Kelsey, Adam Shajnfeld, Adithya Gangidi, Adolfo Victoria, Ahuva Goldstand, Ajay Menon, Ajay Sharma, Alex Boesenberg, Alex Vaughan, Alexei Baevski, Allie Feinstein, Amanda Kallet, Amit Sangani, Anam Yunus, Andrei Lupu, Andres Alvarado, Andrew Caples, Andrew Gu, Andrew Ho, Andrew Poulton, Andrew Ryan, Ankit Ramchandani, Annie Franco, Aparajita Saraf, Arkabandhu Chowdhury, Ashley Gabriel, Ashwin Bharambe, Assaf Eisenman, Azadeh Yazdan, Beau James, Ben Maurer, Benjamin Leonhardi, Bernie Huang, Beth Loyd, Beto~De Paola, Bhargavi Paranjape, Bing Liu, Bo~Wu, Boyu Ni, Braden Hancock, Bram Wasti, Brandon Spence, Brani Stojkovic, Brian Gamido, Britt Montalvo, Carl
  Parker, Carly Burton, Catalina Mejia, Changhan Wang, Changkyu Kim, Chao Zhou, Chester Hu, Ching-Hsiang Chu, Chris Cai, Chris Tindal, Christoph Feichtenhofer, Damon Civin, Dana Beaty, Daniel Kreymer, Daniel Li, Danny Wyatt, David Adkins, David Xu, Davide Testuggine, Delia David, Devi Parikh, Diana Liskovich, Didem Foss, Dingkang Wang, Duc Le, Dustin Holland, Edward Dowling, Eissa Jamil, Elaine Montgomery, Eleonora Presani, Emily Hahn, Emily Wood, Erik Brinkman, Esteban Arcaute, Evan Dunbar, Evan Smothers, Fei Sun, Felix Kreuk, Feng Tian, Firat Ozgenel, Francesco Caggioni, Francisco Guzmán, Frank Kanayet, Frank Seide, Gabriela~Medina Florez, Gabriella Schwarz, Gada Badeer, Georgia Swee, Gil Halpern, Govind Thattai, Grant Herman, Grigory Sizov, Guangyi, Zhang, Guna Lakshminarayanan, Hamid Shojanazeri, Han Zou, Hannah Wang, Hanwen Zha, Haroun Habeeb, Harrison Rudolph, Helen Suk, Henry Aspegren, Hunter Goldman, Ibrahim Damlaj, Igor Molybog, Igor Tufanov, Irina-Elena Veliche, Itai Gat, Jake Weissman, James
  Geboski, James Kohli, Japhet Asher, Jean-Baptiste Gaya, Jeff Marcus, Jeff Tang, Jennifer Chan, Jenny Zhen, Jeremy Reizenstein, Jeremy Teboul, Jessica Zhong, Jian Jin, Jingyi Yang, Joe Cummings, Jon Carvill, Jon Shepard, Jonathan McPhie, Jonathan Torres, Josh Ginsburg, Junjie Wang, Kai Wu, Kam~Hou U, Karan Saxena, Karthik Prasad, Kartikay Khandelwal, Katayoun Zand, Kathy Matosich, Kaushik Veeraraghavan, Kelly Michelena, Keqian Li, Kun Huang, Kunal Chawla, Kushal Lakhotia, Kyle Huang, Lailin Chen, Lakshya Garg, Lavender A, Leandro Silva, Lee Bell, Lei Zhang, Liangpeng Guo, Licheng Yu, Liron Moshkovich, Luca Wehrstedt, Madian Khabsa, Manav Avalani, Manish Bhatt, Maria Tsimpoukelli, Martynas Mankus, Matan Hasson, Matthew Lennie, Matthias Reso, Maxim Groshev, Maxim Naumov, Maya Lathi, Meghan Keneally, Michael~L. Seltzer, Michal Valko, Michelle Restrepo, Mihir Patel, Mik Vyatskov, Mikayel Samvelyan, Mike Clark, Mike Macey, Mike Wang, Miquel~Jubert Hermoso, Mo~Metanat, Mohammad Rastegari, Munish Bansal, Nandhini
  Santhanam, Natascha Parks, Natasha White, Navyata Bawa, Nayan Singhal, Nick Egebo, Nicolas Usunier, Nikolay~Pavlovich Laptev, Ning Dong, Ning Zhang, Norman Cheng, Oleg Chernoguz, Olivia Hart, Omkar Salpekar, Ozlem Kalinli, Parkin Kent, Parth Parekh, Paul Saab, Pavan Balaji, Pedro Rittner, Philip Bontrager, Pierre Roux, Piotr Dollar, Polina Zvyagina, Prashant Ratanchandani, Pritish Yuvraj, Qian Liang, Rachad Alao, Rachel Rodriguez, Rafi Ayub, Raghotham Murthy, Raghu Nayani, Rahul Mitra, Raymond Li, Rebekkah Hogan, Robin Battey, Rocky Wang, Rohan Maheswari, Russ Howes, Ruty Rinott, Sai~Jayesh Bondu, Samyak Datta, Sara Chugh, Sara Hunt, Sargun Dhillon, Sasha Sidorov, Satadru Pan, Saurabh Verma, Seiji Yamamoto, Sharadh Ramaswamy, Shaun Lindsay, Shaun Lindsay, Sheng Feng, Shenghao Lin, Shengxin~Cindy Zha, Shiva Shankar, Shuqiang Zhang, Shuqiang Zhang, Sinong Wang, Sneha Agarwal, Soji Sajuyigbe, Soumith Chintala, Stephanie Max, Stephen Chen, Steve Kehoe, Steve Satterfield, Sudarshan Govindaprasad, Sumit Gupta,
  Sungmin Cho, Sunny Virk, Suraj Subramanian, Sy~Choudhury, Sydney Goldman, Tal Remez, Tamar Glaser, Tamara Best, Thilo Kohler, Thomas Robinson, Tianhe Li, Tianjun Zhang, Tim Matthews, Timothy Chou, Tzook Shaked, Varun Vontimitta, Victoria Ajayi, Victoria Montanez, Vijai Mohan, Vinay~Satish Kumar, Vishal Mangla, Vítor Albiero, Vlad Ionescu, Vlad Poenaru, Vlad~Tiberiu Mihailescu, Vladimir Ivanov, Wei Li, Wenchen Wang, Wenwen Jiang, Wes Bouaziz, Will Constable, Xiaocheng Tang, Xiaofang Wang, Xiaojian Wu, Xiaolan Wang, Xide Xia, Xilun Wu, Xinbo Gao, Yanjun Chen, Ye~Hu, Ye~Jia, Ye~Qi, Yenda Li, Yilin Zhang, Ying Zhang, Yossi Adi, Youngjin Nam, Yu, Wang, Yuchen Hao, Yundi Qian, Yuzi He, Zach Rait, Zachary DeVito, Zef Rosnbrick, Zhaoduo Wen, Zhenyu Yang, and Zhiwei Zhao. 2024.
\newblock \href {https://arxiv.org/abs/2407.21783} {The llama 3 herd of models}.
\newblock \emph{Preprint}, arXiv:2407.21783.

\bibitem[{Eiselen(2016)}]{eiselen2016government}
Roald Eiselen. 2016.
\newblock Government domain named entity recognition for south african languages.
\newblock In \emph{Proceedings of the 10th Language Resources and Evaluation Conference (LREC)}.

\bibitem[{Gage(1994)}]{gage1994new}
Philip Gage. 1994.
\newblock A new algorithm for data compression.
\newblock \emph{The C Users Journal}, 12(2):23--38.

\bibitem[{Gao et~al.(2024)Gao, Tow, Abbasi, Biderman, Black, DiPofi, Foster, Golding, Hsu, Le~Noac'h, Li, McDonell, Muennighoff, Ociepa, Phang, Reynolds, Schoelkopf, Skowron, Sutawika, Tang, Thite, Wang, Wang, and Zou}]{eval-harness}
Leo Gao, Jonathan Tow, Baber Abbasi, Stella Biderman, Sid Black, Anthony DiPofi, Charles Foster, Laurence Golding, Jeffrey Hsu, Alain Le~Noac'h, Haonan Li, Kyle McDonell, Niklas Muennighoff, Chris Ociepa, Jason Phang, Laria Reynolds, Hailey Schoelkopf, Aviya Skowron, Lintang Sutawika, Eric Tang, Anish Thite, Ben Wang, Kevin Wang, and Andy Zou. 2024.
\newblock \href {https://doi.org/10.5281/zenodo.12608602} {A framework for few-shot language model evaluation}.

\bibitem[{Hedderich et~al.(2020)Hedderich, Adelani, Zhu, Alabi, Markus, and Klakow}]{hedderich-etal-2020-transfer}
Michael~A. Hedderich, David Adelani, Dawei Zhu, Jesujoba Alabi, Udia Markus, and Dietrich Klakow. 2020.
\newblock \href {https://doi.org/10.18653/v1/2020.emnlp-main.204} {Transfer learning and distant supervision for multilingual transformer models: A study on {A}frican languages}.
\newblock In \emph{Proceedings of the 2020 Conference on Empirical Methods in Natural Language Processing (EMNLP)}, pages 2580--2591, Online. Association for Computational Linguistics.

\bibitem[{Hedderich et~al.(2021)Hedderich, Lange, Adel, Strötgen, and Klakow}]{hedderich2021surveyrecentapproachesnatural}
Michael~A. Hedderich, Lukas Lange, Heike Adel, Jannik Strötgen, and Dietrich Klakow. 2021.
\newblock \href {https://arxiv.org/abs/2010.12309} {A survey on recent approaches for natural language processing in low-resource scenarios}.
\newblock \emph{Preprint}, arXiv:2010.12309.

\bibitem[{Hendrycks et~al.(2020)Hendrycks, Burns, Basart, Zou, Mazeika, Song, and Steinhardt}]{hendrycks2020measuring}
Dan Hendrycks, Collin Burns, Steven Basart, Andy Zou, Mantas Mazeika, Dawn Song, and Jacob Steinhardt. 2020.
\newblock Measuring massive multitask language understanding.
\newblock \emph{arXiv preprint arXiv:2009.03300}.

\bibitem[{Joshua(2024)}]{joshua2024improving}
Adejumobi~Monjolaoluwa Joshua. 2024.
\newblock \href {https://openreview.net/forum?id=ZTt8f8ILOb} {Improving question-answering capabilities in large language models using retrieval augmented generation ({RAG}): A case study on yoruba culture and language}.
\newblock In \emph{5th Workshop on African Natural Language Processing}.

\bibitem[{Lawal et~al.(2024)Lawal, Adekanmbi, and Soronnadi}]{lawal2024contextual}
Olanrewaju~Israel Lawal, Olubayo Adekanmbi, and Anthony Soronnadi. 2024.
\newblock \href {https://openreview.net/forum?id=NCEnKXztVN} {Contextual evaluation of {LLM}{\textquoteright}s performance on primary education science learning contents in the yoruba language}.
\newblock In \emph{5th Workshop on African Natural Language Processing}.

\bibitem[{Lemerle et~al.(2024)Lemerle, Obin, and Roebel}]{lemerle2024small}
Th{\'e}odor Lemerle, Nicolas Obin, and Axel Roebel. 2024.
\newblock Small-e: Small language model with linear attention for efficient speech synthesis.
\newblock \emph{arXiv preprint arXiv:2406.04467}.

\bibitem[{Lepagnol et~al.(2024)Lepagnol, Gerald, Ghannay, Servan, and Rosset}]{lepagnol2024small}
Pierre Lepagnol, Thomas Gerald, Sahar Ghannay, Christophe Servan, and Sophie Rosset. 2024.
\newblock Small language models are good too: An empirical study of zero-shot classification.
\newblock \emph{arXiv preprint arXiv:2404.11122}.

\bibitem[{Minaee et~al.(2024)Minaee, Mikolov, Nikzad, Chenaghlu, Socher, Amatriain, and Gao}]{minaee2024largelanguagemodelssurvey}
Shervin Minaee, Tomas Mikolov, Narjes Nikzad, Meysam Chenaghlu, Richard Socher, Xavier Amatriain, and Jianfeng Gao. 2024.
\newblock \href {https://arxiv.org/abs/2402.06196} {Large language models: A survey}.
\newblock \emph{Preprint}, arXiv:2402.06196.

\bibitem[{Muhammad et~al.(2023{\natexlab{a}})Muhammad, Abdulmumin, Ayele, Ousidhoum, Adelani, Yimam, Ahmad, Beloucif, Mohammad, Ruder et~al.}]{Muhammad2023AfriSentiAT}
Shamsuddeen Muhammad, Idris Abdulmumin, Abinew Ayele, Nedjma Ousidhoum, David Adelani, Seid Yimam, Ibrahim Ahmad, Meriem Beloucif, Saif Mohammad, Sebastian Ruder, et~al. 2023{\natexlab{a}}.
\newblock Afrisenti: A twitter sentiment analysis benchmark for african languages.
\newblock In \emph{Proceedings of the 2023 Conference on Empirical Methods in Natural Language Processing}, pages 13968--13981.

\bibitem[{Muhammad et~al.(2023{\natexlab{b}})Muhammad, Abdulmumin, Yimam, Adelani, Ahmad, Ousidhoum, Ayele, Mohammad, and Beloucif}]{muhammad2023semeval}
Shamsuddeen~Hassan Muhammad, Idris Abdulmumin, Seid~Muhie Yimam, David~Ifeoluwa Adelani, Ibrahim~Sa'id Ahmad, Nedjma Ousidhoum, Abinew Ayele, Saif~M Mohammad, and Meriem Beloucif. 2023{\natexlab{b}}.
\newblock Semeval-2023 task 12: Sentiment analysis for african languages (afrisenti-semeval).
\newblock \emph{arXiv preprint arXiv:2304.06845}.

\bibitem[{Muhammad et~al.(2022)Muhammad, Adelani, Ruder, Ahmad, Abdulmumin, Bello, Choudhury, Emezue, Abdullahi, Aremu, Jorge, and Brazdil}]{muhammad-etal-2022-naijasenti}
Shamsuddeen~Hassan Muhammad, David~Ifeoluwa Adelani, Sebastian Ruder, Ibrahim~Sa{'}id Ahmad, Idris Abdulmumin, Bello~Shehu Bello, Monojit Choudhury, Chris~Chinenye Emezue, Saheed~Salahudeen Abdullahi, Anuoluwapo Aremu, Al{\'\i}pio Jorge, and Pavel Brazdil. 2022.
\newblock \href {https://aclanthology.org/2022.lrec-1.63} {{N}aija{S}enti: A {N}igerian {T}witter sentiment corpus for multilingual sentiment analysis}.
\newblock In \emph{Proceedings of the Thirteenth Language Resources and Evaluation Conference}, pages 590--602, Marseille, France. European Language Resources Association.

\bibitem[{Nekoto et~al.(2020)Nekoto, Marivate, Matsila, Fasubaa, Kolawole, Fagbohungbe, Akinola, Muhammad, Kabongo, Osei, Freshia, Niyongabo, Macharm, Ogayo, Ahia, Meressa, Adeyemi, Mokgesi{-}Selinga, Okegbemi, Martinus, Tajudeen, Degila, Ogueji, Siminyu, Kreutzer, Webster, Ali, Abbott, Orife, Ezeani, Dangana, Kamper, Elsahar, Duru, Kioko, Murhabazi, Biljon, Whitenack, Onyefuluchi, Emezue, Dossou, Sibanda, Bassey, Olabiyi, Ramkilowan, {\"{O}}ktem, Akinfaderin, and Bashir}]{Nekoto}
Wilhelmina Nekoto, Vukosi Marivate, Tshinondiwa Matsila, Timi~E. Fasubaa, Tajudeen Kolawole, Taiwo Fagbohungbe, Solomon~Oluwole Akinola, Shamsuddeen~Hassan Muhammad, Salomon Kabongo, Salomey Osei, Sackey Freshia, Rubungo~Andre Niyongabo, Ricky Macharm, Perez Ogayo, Orevaoghene Ahia, Musie Meressa, Mofe Adeyemi, Masabata Mokgesi{-}Selinga, Lawrence Okegbemi, Laura~Jane Martinus, Kolawole Tajudeen, Kevin Degila, Kelechi Ogueji, Kathleen Siminyu, Julia Kreutzer, Jason Webster, Jamiil~Toure Ali, Jade~Z. Abbott, Iroro Orife, Ignatius Ezeani, Idris~Abdulkabir Dangana, Herman Kamper, Hady Elsahar, Goodness Duru, Ghollah Kioko, Espoir Murhabazi, Elan~Van Biljon, Daniel Whitenack, Christopher Onyefuluchi, Chris Emezue, Bonaventure Dossou, Blessing~K. Sibanda, Blessing~Itoro Bassey, Ayodele Olabiyi, Arshath Ramkilowan, Alp {\"{O}}ktem, Adewale Akinfaderin, and Abdallah Bashir. 2020.
\newblock \href {https://arxiv.org/abs/2010.02353} {Participatory research for low-resourced machine translation: {A} case study in african languages}.
\newblock \emph{CoRR}, abs/2010.02353.

\bibitem[{Ogundepo et~al.(2023)Ogundepo, Gwadabe, Rivera, Clark, Ruder, Adelani, Dossou, DIOP, Sikasote, Hacheme, Buzaaba, Ezeani, Mabuya, Osei, Emezue, Kahira, Muhammad, Oladipo, Owodunni, Tonja, Shode, Asai, Ajayi, Siro, Arthur, Adeyemi, Ahia, Anuoluwapo, Awosan, Chukwuneke, Opoku, Ayodele, Otiende, Mwase, Sinkala, Rubungo, Ajisafe, Onwuegbuzia, Mbow, Niyomutabazi, Mukonde, Lawan, Ahmad, Alabi, Namukombo, Chinedu, Phiri, Putini, Mngoma, Amuok, Iro, and Adhiambo}]{ogundepo2023afriqa}
Odunayo Ogundepo, Tajuddeen~R. Gwadabe, Clara~E. Rivera, Jonathan~H. Clark, Sebastian Ruder, David~Ifeoluwa Adelani, Bonaventure F.~P. Dossou, Abdou~Aziz DIOP, Claytone Sikasote, Gilles Hacheme, Happy Buzaaba, Ignatius Ezeani, Rooweither Mabuya, Salomey Osei, Chris Emezue, Albert~Njoroge Kahira, Shamsuddeen~H. Muhammad, Akintunde Oladipo, Abraham~Toluwase Owodunni, Atnafu~Lambebo Tonja, Iyanuoluwa Shode, Akari Asai, Tunde~Oluwaseyi Ajayi, Clemencia Siro, Steven Arthur, Mofetoluwa Adeyemi, Orevaoghene Ahia, Aremu Anuoluwapo, Oyinkansola Awosan, Chiamaka Chukwuneke, Bernard Opoku, Awokoya Ayodele, Verrah Otiende, Christine Mwase, Boyd Sinkala, Andre~Niyongabo Rubungo, Daniel~A. Ajisafe, Emeka~Felix Onwuegbuzia, Habib Mbow, Emile Niyomutabazi, Eunice Mukonde, Falalu~Ibrahim Lawan, Ibrahim~Said Ahmad, Jesujoba~O. Alabi, Martin Namukombo, Mbonu Chinedu, Mofya Phiri, Neo Putini, Ndumiso Mngoma, Priscilla~A. Amuok, Ruqayya~Nasir Iro, and Sonia Adhiambo. 2023.
\newblock \href {https://arxiv.org/abs/2305.06897} {Afriqa: Cross-lingual open-retrieval question answering for african languages}.
\newblock \emph{Preprint}, arXiv:2305.06897.

\bibitem[{Orife et~al.(2020)Orife, Kreutzer, Sibanda, Whitenack, Siminyu, Martinus, Ali, Abbott, Marivate, Kabongo, Meressa, Murhabazi, Ahia, Biljon, Ramkilowan, Akinfaderin, {\"{O}}ktem, Akin, Kioko, Degila, Kamper, Dossou, Emezue, Ogueji, and Bashir}]{masakhane}
Iroro Orife, Julia Kreutzer, Blessing~K. Sibanda, Daniel Whitenack, Kathleen Siminyu, Laura Martinus, Jamiil~Toure Ali, Jade~Z. Abbott, Vukosi Marivate, Salomon Kabongo, Musie Meressa, Espoir Murhabazi, Orevaoghene Ahia, Elan~Van Biljon, Arshath Ramkilowan, Adewale Akinfaderin, Alp {\"{O}}ktem, Wole Akin, Ghollah Kioko, Kevin Degila, Herman Kamper, Bonaventure Dossou, Chris Emezue, Kelechi Ogueji, and Abdallah Bashir. 2020.
\newblock \href {https://arxiv.org/abs/2003.11529} {Masakhane - machine translation for africa}.
\newblock In \emph{1st AfricaNLP Workshop Proceedings, AfricaNLP@ICLR 2020, Virtual Conference, Formerly Addis Ababa Ethiopia, 26th April 2020}.

\bibitem[{Papineni et~al.(2002)Papineni, Roukos, Ward, and Zhu}]{papineni2002bleu}
Kishore Papineni, Salim Roukos, Todd Ward, and Wei-Jing Zhu. 2002.
\newblock Bleu: a method for automatic evaluation of machine translation.
\newblock In \emph{Proceedings of the 40th annual meeting of the Association for Computational Linguistics}, pages 311--318.

\bibitem[{Qin et~al.(2024)Qin, Chen, Zhou, Chen, Li, Liao, Li, Che, and Yu}]{qin2024multilinguallargelanguagemodel}
Libo Qin, Qiguang Chen, Yuhang Zhou, Zhi Chen, Yinghui Li, Lizi Liao, Min Li, Wanxiang Che, and Philip~S. Yu. 2024.
\newblock \href {https://arxiv.org/abs/2404.04925} {Multilingual large language model: A survey of resources, taxonomy and frontiers}.
\newblock \emph{Preprint}, arXiv:2404.04925.

\bibitem[{Sanyal et~al.(2024)Sanyal, Sanghavi, and Dimakis}]{sanyal2024pre}
Sunny Sanyal, Sujay Sanghavi, and Alexandros~G Dimakis. 2024.
\newblock Pre-training small base lms with fewer tokens.
\newblock \emph{arXiv preprint arXiv:2404.08634}.

\bibitem[{Scaria et~al.(2024)Scaria, Kennedy, and Subramani}]{scaria2024can}
Nicy Scaria, Silvester John~Joseph Kennedy, and Deepak Subramani. 2024.
\newblock Can small language models learn, unlearn, and retain noise patterns?
\newblock \emph{arXiv preprint arXiv:2407.00996}.

\bibitem[{Song et~al.(2024)Song, Huang, Yin, Yang, and Gao}]{song2024achieving}
Jifeng Song, Kai Huang, Xiangyu Yin, Boyuan Yang, and Wei Gao. 2024.
\newblock Achieving sparse activation in small language models.
\newblock \emph{arXiv preprint arXiv:2406.06562}.

\bibitem[{Team et~al.(2024)Team, Mesnard, Hardin, Dadashi, Bhupatiraju, Pathak, Sifre, Rivi{\`e}re, Kale, Love et~al.}]{gemma_2024}
Gemma Team, Thomas Mesnard, Cassidy Hardin, Robert Dadashi, Surya Bhupatiraju, Shreya Pathak, Laurent Sifre, Morgane Rivi{\`e}re, Mihir~Sanjay Kale, Juliette Love, et~al. 2024.
\newblock Gemma: Open models based on gemini research and technology.
\newblock \emph{arXiv preprint arXiv:2403.08295}.

\bibitem[{Team et~al.(2022)Team, Costa-jussà, Cross, Çelebi, Elbayad, Heafield, Heffernan, Kalbassi, Lam, Licht, Maillard, Sun, Wang, Wenzek, Youngblood, Akula, Barrault, Gonzalez, Hansanti, Hoffman, Jarrett, Sadagopan, Rowe, Spruit, Tran, Andrews, Ayan, Bhosale, Edunov, Fan, Gao, Goswami, Guzmán, Koehn, Mourachko, Ropers, Saleem, Schwenk, and Wang}]{nllbteam2022languageleftbehindscaling}
NLLB Team, Marta~R. Costa-jussà, James Cross, Onur Çelebi, Maha Elbayad, Kenneth Heafield, Kevin Heffernan, Elahe Kalbassi, Janice Lam, Daniel Licht, Jean Maillard, Anna Sun, Skyler Wang, Guillaume Wenzek, Al~Youngblood, Bapi Akula, Loic Barrault, Gabriel~Mejia Gonzalez, Prangthip Hansanti, John Hoffman, Semarley Jarrett, Kaushik~Ram Sadagopan, Dirk Rowe, Shannon Spruit, Chau Tran, Pierre Andrews, Necip~Fazil Ayan, Shruti Bhosale, Sergey Edunov, Angela Fan, Cynthia Gao, Vedanuj Goswami, Francisco Guzmán, Philipp Koehn, Alexandre Mourachko, Christophe Ropers, Safiyyah Saleem, Holger Schwenk, and Jeff Wang. 2022.
\newblock \href {https://arxiv.org/abs/2207.04672} {No language left behind: Scaling human-centered machine translation}.
\newblock \emph{Preprint}, arXiv:2207.04672.

\bibitem[{Thawakar et~al.(2024)Thawakar, Vayani, Khan, Cholakal, Anwer, Felsberg, Baldwin, Xing, and Khan}]{thawakar2024mobillama}
Omkar Thawakar, Ashmal Vayani, Salman Khan, Hisham Cholakal, Rao~M Anwer, Michael Felsberg, Tim Baldwin, Eric~P Xing, and Fahad~Shahbaz Khan. 2024.
\newblock Mobillama: Towards accurate and lightweight fully transparent gpt.
\newblock \emph{arXiv preprint arXiv:2402.16840}.

\bibitem[{Xu et~al.(2024)Xu, Han, Yang, Wang, Zhu, Liu, Liu, and Che}]{xu2024onebit}
Yuzhuang Xu, Xu~Han, Zonghan Yang, Shuo Wang, Qingfu Zhu, Zhiyuan Liu, Weidong Liu, and Wanxiang Che. 2024.
\newblock Onebit: Towards extremely low-bit large language models.
\newblock \emph{arXiv preprint arXiv:2402.11295}.

\bibitem[{Zhang et~al.(2024)Zhang, Zeng, Wang, and Lu}]{zhang2024tinyllama}
Peiyuan Zhang, Guangtao Zeng, Tianduo Wang, and Wei Lu. 2024.
\newblock Tinyllama: An open-source small language model.
\newblock \emph{arXiv preprint arXiv:2401.02385}.

\bibitem[{Zhu et~al.(2024)Zhu, Zhu, Liu, Ou, Mou, and Tang}]{zhu2024llava}
Yichen Zhu, Minjie Zhu, Ning Liu, Zhicai Ou, Xiaofeng Mou, and Jian Tang. 2024.
\newblock Llava-phi: Efficient multi-modal assistant with small language model.
\newblock \emph{arXiv preprint arXiv:2401.02330}.

\end{thebibliography}

\appendix

% \section{Example Appendix}
% \label{sec:appendix}

% This is an appendix.

\end{document}